\documentclass[letterpaper]{article}
\usepackage[margin=1in]{geometry}

\usepackage{natbib}
\PassOptionsToPackage{numbers, compress}{natbib}
\usepackage[utf8]{inputenc} % allow utf-8 input
\usepackage[T1]{fontenc}    % use 8-bit T1 fonts
\usepackage{lmodern}
\usepackage{hyperref}       % hyperlinks
\usepackage{url}            % simple URL typesetting
\usepackage{booktabs}       % professional-quality tables
\usepackage{amsfonts}       % blackboard math symbols
\usepackage{nicefrac}       % compact symbols for 1/2, etc.
\usepackage{microtype}      % microtypography
\usepackage{xcolor}         % colors

\usepackage{amsmath}
\usepackage{amssymb}
\usepackage{mathtools}
\usepackage{amsthm}

\usepackage{amsmath,amsfonts,bm, amsthm}
\usepackage{thmtools}  
\usepackage{thm-restate}

\newcommand{\R}{\mathbb{R}}
\newcommand{\E}{\mathop{\mathbb{E}}}
\newcommand{\Ex}{\mathbb{E}}
\newcommand{\argmin}{\mathop{\text{argmin}}}

\newcommand{\bx}{\mathbf{x}}

\newcommand{\by}{\mathbf{y}}
\usepackage{mathtools}

\newcommand{\textmeasure}[1]{\textrm{#1}} % set the text font for properties
\newcommand{\gdx}{\textmeasure{update\_corr}}
\newcommand{\gdelta}{\textmeasure{update\_corr\_RS}}
\newcommand{\dloss}{\textmeasure{loss\_diff}}
\newcommand{\instGap}{\textmeasure{inst\_gap}}
\newcommand{\avgGap}{\textmeasure{avg\_gap}}
\newcommand{\expGap}{\textmeasure{exp\_gap}}
\newcommand{\convRatio}{\textmeasure{convexity\_ratio}}
\newcommand{\instSmooth}{\textmeasure{inst\_smooth}}
\newcommand{\expSmooth}{\textmeasure{exp\_smooth}}

\newcommand{\maxSmooth}{\textmeasure{max\_smooth}}
\def\eqref#1{equation~\ref{#1}}
\def\1{\bm{1}}

\DeclareMathAlphabet{\mathsfit}{\encodingdefault}{\sfdefault}{m}{sl}
\SetMathAlphabet{\mathsfit}{bold}{\encodingdefault}{\sfdefault}{bx}{n}

\usepackage{wrapfig}
\usepackage{float}
\usepackage{subfigure}
\usepackage[font=small,labelfont=bf]{caption}  % optional: adjust caption style
\usepackage[capitalise]{cleveref}

\theoremstyle{plain}

\theoremstyle{definition}

\theoremstyle{remark}

\title{Reevaluating Theoretical Analysis Methods for Optimization in Deep Learning}

\author{
  Hoang Tran \\
  Boston University \\
  \texttt{tranhp@bu.edu} \\
  \and
  Qinzi Zhang \\
  Boston University \\
  \texttt{qinziz@bu.edu} \\
  \and
  Ashok Cutkosky \\
  Boston University \\
  \texttt{cutkosky@bu.edu} \\
}

\begin{document}

\maketitle

\begin{abstract}
There is a significant gap between our theoretical understanding of optimization algorithms used in deep learning and their practical performance. Theoretical development usually focuses on proving convergence guarantees under a variety of different assumptions, which are themselves often chosen based on a rough combination of intuitive match to practice and analytical convenience. In this paper, we carefully measure the degree to which the standard optimization analyses are capable of explaining modern algorithms. To do this, we develop new empirical metrics that compare real optimization behavior with analytically predicted behavior. Our investigation is notable for its tight integration with modern optimization analysis: rather than simply checking high-level assumptions made in the analysis (e.g. smoothness), we also verify key low-level identities used by the analysis to explain optimization behavior that might hold even if the high-level motivating assumptions do not. Notably, we find that smoothness‑based analyses fail in practice under most scenarios,
% ---both the smoothness conditions and the corresponding key identities. On the other hand, 
but the key identities commonly used in convex‑optimization analyses often hold in practice despite the objective’s global non‑convexity.
\end{abstract}

\section{Introduction}
In optimization theory, algorithmic development and analysis requires a set of assumptions about the functions we aim to optimize. These assumptions fundamentally influence the behavior of optimization algorithms and their efficacy in practice.  For example, Adagrad \citep{JMLR:v12:duchi11a,mcmahan2010adaptive} (which later inspired Adam \citep{kingma2014adam}) classically relies on the convexity assumption to provide a theoretical convergence guarantee.
When the loss is non-convex, a variety of alternate assumptions are deployed, such as smoothness (e.g. a bounded Hessian) \citep{ghadimi2013stochastic,carmon2017convex,li2019convergence,ward2020adagrad,wang2023convergence} or ``weak convexity'' \citep{davis2019stochastic,mai2020convergence,liu2023convergence}. 
% Similarly, other recent developments in non-convex optimization based on variance-reduction such as SVRG \citep{NIPS2013_ac1dd209}, SARAH \citep{nguyen2017sarah}, or STORM \citep{cutkosky2019momentum} rely heavily on the smoothness assumption of the objective functions to control the difference between the estimated gradient and the gradient evaluated at some referenced points. 
If these conditions are not met, the convergence analyses of these algorithms may longer hold. In this paper, we systematically verify these assumptions and related optimization analyses across various deep learning tasks using simple, computationally feasible methods. We hope that our findings will serve as a guideline for future research, helping to develop theoretical frameworks that are both analytically tractable and practically applicable.

% Despite their critical importance, these assumptions are often overlooked in practical applications due to the challenges and time-consuming nature of verifying such global conditions. 
% Mathematical studies frequently assume the validity of these assumptions in practice or only examine conditions relevant to specific algorithms or problem settings. 

Importantly, we do not want to simply ask ``do current assumptions apply to deep neural networks''. Instead, we wish to ask whether the \emph{analyses} based on currently prevalent techniques can predict current practical performance. This is a subtly different question: it turns out that most modern analyses actually rely on a few key identities. These identities are usually \emph{empirically measurable} from the iterates of an optimizer. In theoretical analysis, these identities are controlled via various global assumptions (such as convexity or smoothness), but we instead measure directly these identities. This has a significant advantage: not only can it falsify the global assumptions, it can tell if any \emph{different} assumptions can be made that would ``rescue'' the analysis.

\paragraph{Our Contributions}
We propose simple, on-the-fly measures that capture how well modern analyses describe practice, and we measure them on a wide range of tasks, including basic convex optimization problems, image classification tasks using deep residual networks, and large language models (LLMs) pre-training. 
In Section \ref{sec:convex}, we introduce convexity gap and convexity ratio to capture key identities in convexity‑based analyses. In Section \ref{sec:smoothness}, we introduce update correlation for smoothness-based analyses.
Of independent interest, we also develop a new smoothness measure that closely approximates the sharpness of a function. The new measure is computationally feasible even for very deep networks, where computing the standard sharpness measure via hessian becomes expensive. 
% This allows for the use of our measure in studying flat/sharp minima and their implications for generalization in much larger networks. 
% Finally, we offer alternative theoretical analyses for cases where common theoretical assumptions do not hold.

Our experiment results suggest that although the objective is globally non-convex, surprisingly, a key quantity in almost all convexity-based analyses, that we call the ``convexity ratio'', is typically positive. A positive ratio does \emph{not} require actual convexity, but \emph{does} imply that convexity-based analyses may nevertheless inform real optimization performance. This observation helps explain the empirical success of optimizers which stem from online convex optimization, such as AdaGrad \citep{duchi10adagrad, mcmahan2010adaptive}, Adam \citep{kingma2014adam}, Shampoo \citep{gupta2018shampoo}, and etc. Furthermore, we observe that the objective is non-smooth neither globally nor locally along the training trajectory, and our measures of update correlation suggests that the analytical techniques central to smooth optimization often do not hold in practice.
% Our results suggest that most analytical techniques do \emph{not} describe practical performance. Our work fits into a recent trend of challenging and moving past classical optimization assumptions  \cite{simsekli2019tail, zhang2020complexity, zhang2020adaptive, davis2021gradient, davis2020stochastic}. However, our focus is not on algorithm development. Instead, we simply want to promote empirical verification of  optimization analysis.

We highlight that our focus is neither on algorithm development nor on proposing new assumptions. Instead, we simply want to promote empirical verification of optimization analyses. Overall, we feel that our findings motivate two actions in the research community: first, it is important to develop new assumptions and analytical techniques to understand modern optimization. Second, we advocate for verifying any new assumptions by carefully measuring quantities that \emph{actually appear in the optimization analysis} rather than attempting to verify global assumptions.
\section{Related Works}

% In the theoretical study of stochastic optimization, convexity and smoothness have been two of the most important properties. As a general overview, the development of theoretical work began with understanding convex problems, with SGD as the initial algorithm of choice \cite{bottou08large}. Subsequent works utilized convexity to propose various algorithms, most famously including AdaGrad \citep{duchi10adagrad, mcmahan2010adaptive}. When the objective is non-convex, various forms of smoothness have long been necessary assumptions, including first-order smoothness \citep{ghadimi2013stochastic, carmon2017convex}, second-order smoothness \citep{tripuraneni2018stochastic, carmon2018accelerated, fang2019sharp, arjevani2020second}, and mean-square smoothness \citep{zhou2018stochastic, fang2018spider, cutkosky2019momentum}. For objectives that are neither convex nor smooth, some works have studied the Moreau envelope as a smoothed version of weakly convex objectives \citep{davis2019stochastic, mai2020convergence}. For objectives that are not even weakly convex, recent studies have studied Goldstein stationary point as alternative convergence criterion \citep{zhang2020complexity, cutkosky2023optimal}. 
% % We will discuss each sub-field in details in later sections.

There have been extensive studies on the empirical properties and the loss landscape of modern machine learning. \citet{goodfellow2015qualitative} proposed one-dimensional and two-dimensional visualization tools for the loss landscape of various neural networks, demonstrating that SGD rarely encounters local minima during training. \citet{im2017empirical} tested the training trajectories of different optimizers using the same visualization tools and observed that different optimizers exhibit distinct behaviors when encountering saddle points. \citet{li2018visualizing} proposed more refined visualization techniques and showed that the smoothness of the loss landscape closely correlates with generalization performance. 
% \citet{nakkiran2019sgd} studied the dynamics of SGD training, showing that SGD learns simple classifiers at early training stages and learns more complex classifiers at later stages.
\citet{power2022grokking} reported the grokking phenomenon on a synthesized dataset such that after a long period of severe overfitting, validation score suddenly increases to almost perfect generalization.
\citet{thilak2022slingshot} revealed the slingshot effect of training neural networks with adaptive optimizers, which is a cyclic behavior between stable and unstable regimes during training process.
% \citet{belkin2019reconciling, nakkiran2019deep} demonstrated the double-descent effect where increasing model complexity in the over-parameterized regime effectively decreases the generalization loss. 
While these results provide general insight into neural network landscapes, we focus on validating common assumptions and key identities fundamental to the analysis of optimization theory.

There are several studies that align more closely with our work. \citet{xing2018walk} demonstrated that loss interpolation between consecutive iterates is locally convex, which agrees with our observations in Sec \ref{subsec:localconvex}. While their experiments focus on SGD and image classification tasks, we extended the scope of our convexity measures to include AdamW and LLMs. Furthermore, we also tested a more global convexity measure in Sec \ref{subsec:global-convex}.
\citet{cohen2020gradient, cohen2022adaptive} observed the ``edge of stability'' phenomenon where the sharpness increases during early stage of training and then stabilizes. Our observations in Sec \ref{sec:smoothness} align with their finding and extend beyond CIFAR-10 tasks. 
% Moreover, we proposed a smoothness measure that achieves the same goal of measuring sharpness but is computationally simpler, facilitating sharpness measurements for complex models like LLMs.
\citet{rosenfeld2023outliers} demonstrated the opposing signal phenomenon that there are groups of outliers such that decreasing loss over one group increases loss over other groups, which could explain our observation of positive update correlation in Sec \ref{subsec:update-corr}. 
Unlike these works, our work does not only verify common assumptions but also directly measures key quantities in modern analyses.

\section{Background and Experiment Setup}
In typical optimization analysis for machine learning, the goal is to minimize an objective $F$ given by
$F(\bx) = \Ex_{z \sim P_z}[f(\bx, z)],$
% \begin{align*}
%     F(\bx) = \Ex_{z \sim P_z}[f(\bx, z)],
% \end{align*}
where $f(\bx, z): \R^d \times \mathcal{Z} \mapsto \R$ is a differentiable function of $\bx \in \R^d$. $\bx$ indicates the model parameters, $z \in \mathcal{Z}$ indicates an example data point or minibatch of examples, and $P_z$ is some data distribution. 
% The function $F$ represents either a train loss or a population loss depending on various details of the problem setup.

The most common paradigm in optimization analysis is the following three-step strategy: first, identify a "convergence criterion" of interest - for example the loss of some weights output by an algorithm minus the loss of the optimal weights. Second, identify an algebraic expression that can be related to this convergence criterion (often through use of some assumption on the loss landscape). Finally, establish that a given algorithm can guarantee a bound on this algebraic expression (often again using some assumption on the loss landscape):
\begin{align}
   \underbrace{\text{Convergence Criterion}}_{\text{e.g. }\frac{1}{T}\sum_{t=1}^T F(\bx_t) - F(\bx_\star) }
   &\le \underbrace{\text{Algebraic Expression}}_{\text{e.g. }\frac{1}{T}\sum_{t=1}^T \langle \nabla F(\bx_t), \bx_t - \bx_\star\rangle} 
   \le \underbrace{\text{Upper Bound}}_{\text{e.g. }O(1/\sqrt{T})}\label{eqn:paradigm}
\end{align}
The example above summarizes the analysis of SGD for convex objectives, in which $\bx_\star =\argmin F$, and the middle ``algebraic expression'' is often termed the \emph{regret} (see \cite{orabona2019modern, hazan2022introduction} for details).

This paradigm is used in two different ways: first, from a \emph{scientific} perspective, one can try to prove convergence properties for well-known algorithms such as AdamW to explain why these algorithms work well in practice (see e.g. \cite{li2019convergence, faw2022power, ward2020adagrad, zaheer2018adaptive, reddi2019convergence}). Second, from an \emph{engineering} perspective, one can try to design better optimizers from first principles. For this second use-case, the typical approach is to identify a large class algorithms, such as SGD parametrized by the learning rate, and then choose the member of this class that analytically minimizes the upper bound (see e.g. \cite{duchi10adagrad, mcmahan2010adaptive, hazan2007adaptive, ghadimi2013stochastic}). This exact approach is how the AdaGrad family of algorithms (which was the intellectual precursor to Adam) was developed.

In order for this paradigm to provide meaningful answers, we must believe that the inequalities in equation (\ref{eqn:paradigm}) hold at least approximately. We can investigate this from two angles: first, we can ask whether the original assumptions that motivated the analysis hold. Second, we can often \emph{empirically measure} expressions related to those appearing in (\ref{eqn:paradigm}), and check the degree to which the desired inequalities hold. These are more likely to hold than the underlying assumptions, because the assumptions imply the inequalities, but the reverse may not be true.

Empirical verification of these inequalities is made especially attractive for two reasons. First, many optimization analyses actually use only a few options for the ``algebraic expression'' in (\ref{eqn:paradigm}): the only thing that changes is the analysis of the algorithm leading to improved upper bounds. Thus, by empirically measuring the degree to which the \emph{first} inequality in (\ref{eqn:paradigm}) holds, we can interrogate whether popular analyses strategies can explain optimization success in deep learning in way that is less tightly coupled to whether particular global assumptions hold or not.

Two very popular assumptions about the loss landscape and the optimization process are smoothness and convexity.
Formally, a differentiable function $f(\cdot, \cdot):  \R^d \times \mathcal{Z} \mapsto \R$ is convex if it satisfies:
\begin{align*}
    f(\by,z) &\ge f(\bx,z) + \langle \nabla f(\bx,z),\by-\bx\rangle,  \quad \forall \bx,\by \in \R^d, z \in \mathcal{Z}.
\end{align*}
Further, $f(\cdot, \cdot)$ is $L-$smooth if it satisfies:
\begin{align*}
    \|\nabla f(\bx,z) - \nabla f(\by,z)\|&\le L\|\bx-\by\|,  \quad \forall \bx,\by \in \R^d, z \in \mathcal{Z}.
\end{align*}
These are some of the most common assumptions in optimization theory \citep{zinkevich2003online,duchi10adagrad,ghadimi2013stochastic,bubeck2015convex,carmon2017convex,NEURIPS2020_93931410,NEURIPS2019_21ce6891,hazan2022introduction,cutkosky2019momentum}. We would like to quantify them in our experiments. Computing the global smoothness constant as well as the convexity of the true loss functions $F(\bx)$ is infeasible. Fortunately, typical optimization analysis \emph{does not actually require} these properties to hold globally. We instead measure proxies that we call the \textit{instantaneous convexity gap}, denoted by {\instGap}, and \textit{instantaneous smoothness}, denoted by {\instSmooth}, to estimate the levels of convexity and smoothness of the true loss function. These proxies are more closely related to the expressions that arise in typical analysis than unobservable global values. Formally, the $\instGap$ with respect to a reference point $\by_t$ is as follows: 
% $ \instGap_t(\by_t) \coloneqq f(\bx_t,z_t) - f(\by_t,z_t) - \langle \nabla f(\bx_t,z_t), \bx_t - \by_t\rangle $.
\begin{align}
    % \text{inst$\_$gap}_t(\by) 
    \instGap_t(\by_t) 
    &\coloneqq f(\bx_t,z_t) - f(\by_t,z_t) - \langle \nabla f(\bx_t,z_t), \bx_t - \by_t\rangle.
    \label{eq:convexityGap}
\end{align}
In our measurements, we use two settings for $\by_t$. First, we consider $\by_t =\bx_{t-1}$ to analyze the properties of consecutive points and their impact on the optimization path. Next, we use the constant value $\by_t = \bx^\star$, where $\bx^\star$ is the \emph{final} iterate produced by a either the same training run or a training run with a different random seed. This setting provides a more global view of the loss landscape. If $f$ is convex, 
% $f(\by_t,z_t) \ge f(\bx_t,z_t) + \langle \nabla f(\bx_t,z_t), \by_t - \bx_t\rangle $ and 
then $\instGap_t$ should be non-positive (but the reverse may not hold). We also compute the average convexity gap and the exponential moving average of the convexity gaps with respect to a sequence of reference points $\by_1,\ldots,\by_t$ (denoted as $\by_{1:t}$ for short), respectively defined as 
\begin{align}
    \textstyle \avgGap_t(\by_{1:t}) &= \textstyle\frac{1}{t}\sum_{i=1}^t \instGap_i(\by_i), \notag\\
    \expGap_t(\by_{1:t}) &= \beta \cdot \expGap_{t-1}(\by_{1:t-1})
    + (1-\beta) \cdot \instGap_t(\by_t).
    \label{eq:avg,exp-conv_gap}
\end{align}
where $\beta\in(0,1)$ (we choose $\beta=0.99$ for our measurements).

Next, we define the instantaneous smoothness at iteration $t$ with respect to $\by_t$ as:
\begin{align}
    \textstyle
    \instSmooth_t(\by_t)  = \frac{\|\nabla f(\bx_t,z_t) - \nabla f(\by_t,z_t)\|}{\|\bx_t - \by_t\|}
    \label{eq:inst_smooth}
\end{align}
If the loss function is $L$-smooth, then $\instSmooth_t\le L$ for all $t\in [T]$. Thus, if this instantaneous smoothness quantity is uniformly bounded by a constant, it could indicate that our loss landscape is smooth. Similar to the convexity gap, we also keep track of other forms of the smoothness measure such as the maximum smoothness 
% (the highest value observed across all iterations) 
and the exponential average smoothness, respectively defined as
\begin{align}
    \maxSmooth_t(\by_{1:t}) &= \max_{i\le t} \instSmooth_i(\by_i), \notag\\
    \expSmooth_t(\by_{1:t}) &= \beta \cdot \expSmooth_{t-1}(\by_{1:t-1}) 
    + (1-\beta)\cdot \instSmooth_t(\by_t).
    \label{eq:avg_exp-smooth}
\end{align}
Our maximum smoothness is similar to the smoothness metric proposed in \citep{santurkar2018does,zhang2019gradient}. However, instead of tracking the largest smoothness value along the line of the update difference $\bx_t-\bx_{t-1}$, we keep track of the largest value across all iterations.

We conduct experiments across a diverse array of tasks, ranging from simple convex problems to training transformer-based language models. For convex tasks, we run gradient descent on a synthetic dataset using squared loss and also perform logistic regression on various OpenML datasets (Aloi, Connect-4, Covertype, Poker). In the realm of non-convex tasks, we address both Image Classification and NLP benchmarks. For Image Classification tasks, we train popular benchmark datasets Cifar10 and Imagenet \citep{deng2009imagenet} on Resnet18 \citep{he2016deep} using SGD with momentum (SGDM) and AdamW. We use the configurations reported in \citep{yao2020adahessian,tran2022better}. For NLP tasks, we pre-train Bert \citep{devlin2018bert} using the C4 dataset \citep{2019t5} and GPT2 \citep{radford2019language} using the Pile dataset \citep{pile}. Both tasks are trained using SGDM and AdamW. 
% The learning rates for each optimizer are fine-tuned through a grid search in the range $[10^{-6},0.1]$. We also run experiments with Diffusion Model 

% For NLP tasks, we report the metrics computed at checkpoints every 10,000 iterations. For all other experiments, we report the metrics computed at the end of every epoch.
% Further details about the experiment setup can be found in Appendix \ref{subsec:expConfig}.

% To further understand the optimization process of Machine Learning tasks, we also track other quantities such as gradient norm, gradient correlation, progress of the gradient variances, etc.

Most of our training runs involve multiple epochs. In this case, for the non-instantaneous metrics, we ``reset'' the averages at the start of each epoch so that the averages contain only iterates from the current epoch. For the LLM pre-training tasks, due to the large size of the datasets used in these tasks, we completed the training without traversing the entire dataset. Hence, we do not reset our metrics in these experiments. Besides smoothness and convexity, we also track other less-important properties describing the training trajectory. We defer these results to the Appendix due to space limit.

% These metrics collectively offer deeper insights into the dynamic behavior of the loss function throughout the optimization process.
\section{Measuring Convexity}\label{sec:convex}

Convexity is a fundamental assumption in optimization theory since convex functions have many pleasant theoretical guarantees. For instance, every local minimum of a convex function is also a global minimum, which allows us to derive bounds on the suboptimality gap \citep{NIPS2007_0d3180d6,defazio2014saga,cutkosky2019anytime}. Unfortunately, the landscape of deep learning training is known to be non-convex \citep{jain2017non,li2018visualizing,garipov2018loss,choromanska2015loss} due to the complex architectures of deep learning models and the nonlinearity of the activation functions. However, the degree of non-convexity in practical scenarios still remains a bit of a mystery. In this section, we aim to quantify the level of convexity across various machine learning tasks. 
As a sanity check, we first examine the instantaneous convexity gaps with respect to the previous iterate in convex tasks to verify that they align with our theoretical expectations.

% Single Column
% \begin{wrapfigure}{r}{0.5\textwidth}
%     \centering
%     \vspace{-1em}
%     \includegraphics[width=\linewidth]{imgs/inst_gap_convex_new.png}
%     \caption{Instantaneous convexity gap w.r.t. $\bx_{t-1}$ of GD on squared loss (left) and Logistic Regression with OpenML datasets \citep{OpenML2013} (right).}
%     \label{fig:convex_gap}
%     \vspace{-1em}
% \end{wrapfigure}

% Single Column (no wraparound)
\begin{figure}[h]
    \centering
    \includegraphics[width=0.5\linewidth]{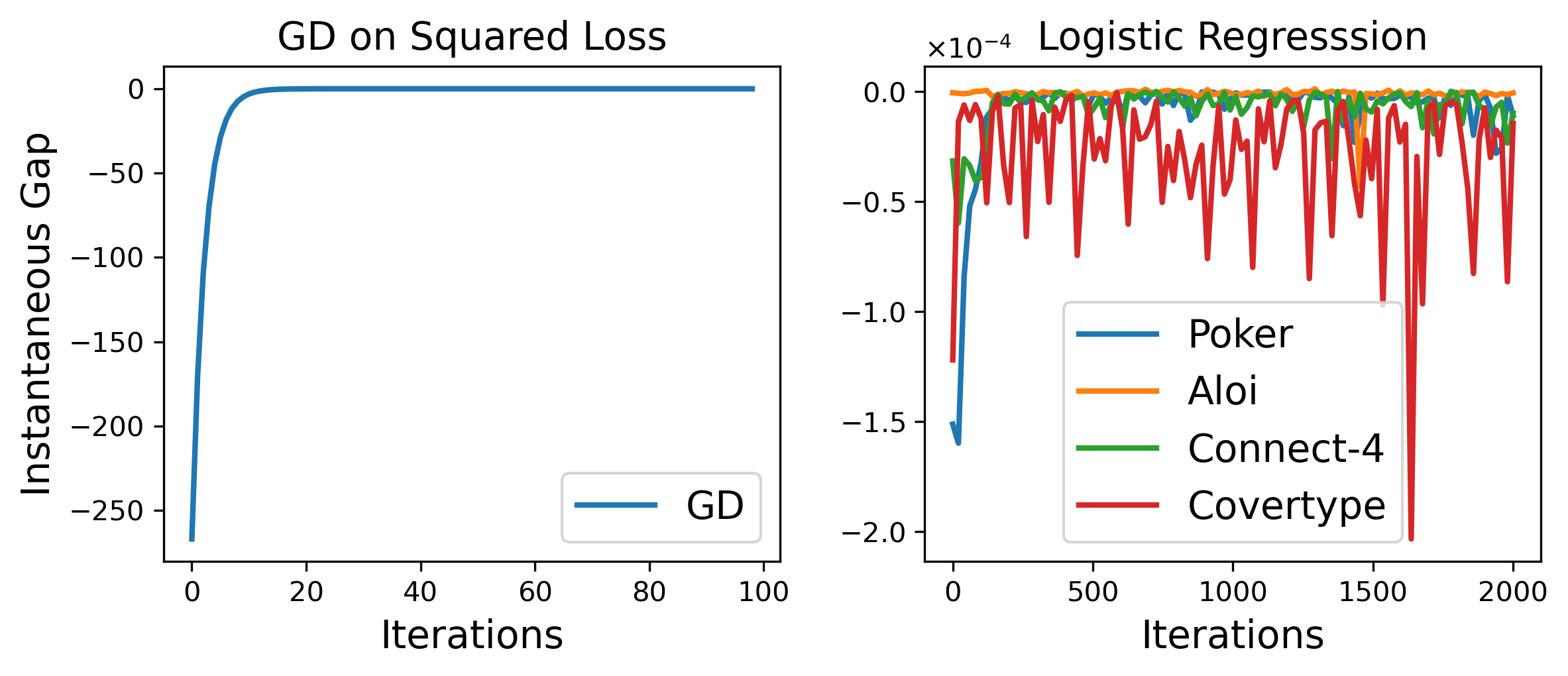}
    \caption{Instantaneous convexity gap w.r.t. $\bx_{t-1}$ of GD on squared loss (left) and Logistic Regression with OpenML datasets \citep{OpenML2013} (right).}
    \label{fig:convex_gap}
\end{figure}

% Double Column
% \begin{figure}[ht]
%     \centering
%     \includegraphics[width=\linewidth]{imgs/inst_gap_convex_new.png}
%     \vspace{-1em}
%     \caption{Instantaneous convexity gap w.r.t. $\bx_{t-1}$ of GD on squared loss (left) and Logistic Regression with OpenML datasets \citep{OpenML2013} (right).}
%     \label{fig:convex_gap}
% \end{figure}

As in Fig.\ref{fig:convex_gap}, the convexity gap is always non-positive, implying global non-convexity of the objective as expected. Now, let us turn our attention to more complex deep learning tasks.

\subsection{Are Deep Learning Loss Landscapes Convex Along Optimization Paths?}
\label{subsec:localconvex}

In this section, we aim to examine the convexity along the paths taken by two popular optimizers, AdamW and SGDM. In particular, we compute both the average and the exponential average convexity gaps, as defined in (\ref{eq:avg,exp-conv_gap}), across various deep learning tasks. Setting $\by_t=\bx_{t-1}$ allows us to measure local convexity in a small step along the optimization trajectory. The presence of any positive gap would imply global non-convexity, and the average of convexity gaps indicates whether an optimization trajectory is ``sufficiently'' convex, i.e. whether instantaneous non-convexity is a rare event. Since stochastic optimization analysis typically involves taking the summation or average from the convexity inequality, one might hope that standard analyses would still hold when the objective is convex in average, i.e., if the average convexity gap is non-positive.
% in expectation. Typical optimization analyses do not actually require instaneous convexity, If the average gap indicates convexity, convex analysis can still be applied to derive convergence rates for many algorithms. 
 % We also provide the instantaneous gap results in Section \ref{sec:extraExp} in the Appendix.
% Additionally, we compute the exponential average of convexity gaps. With $\beta=0.99$, it offers a more localized perspective by averaging over approximately 100 iterations. Results of instantaneous convexity gaps are presented in Fig.\ref{fig:instgap}, and the results of the averages are presented in Fig.\ref{fig:avggap}.

% Single Column
% \begin{wrapfigure}{r}{0.5\textwidth}
%     \centering 
%     \vspace{-1em}
%     % \includegraphics[width=\linewidth]{imgs/convex_gap_new.png}
%     \includegraphics[width=\linewidth]{imgs/convexity_gap_average_2x2.png}
%     \caption{Average and exponential average convexity gap w.r.t. $\by_t=\bx_{t-1}$. A negative value implies that the objective is locally convex in average.}
%     \label{fig:avggap}
%     \vspace{-2em}
% \end{wrapfigure}

% Single Column
\begin{figure}[h]
    \centering 
    \includegraphics[width=0.5\linewidth]{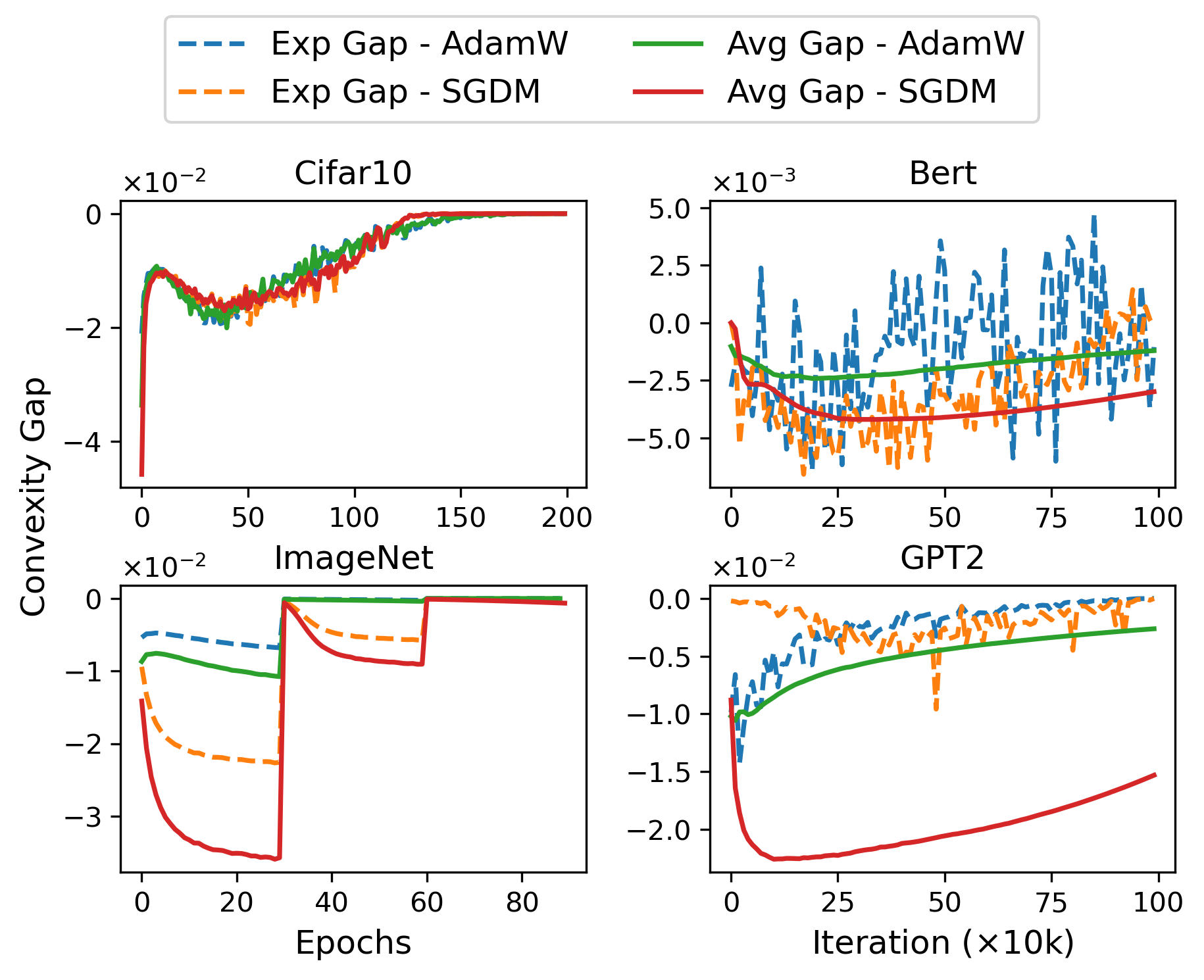}
    \caption{Average and exponential average convexity gap w.r.t. $\by_t=\bx_{t-1}$. A negative value implies that the objective is locally convex in average.}
    \label{fig:avggap}
\end{figure}

% Double Column
% \begin{figure}[ht]
%     \centering 
%     % \includegraphics[width=\linewidth]{imgs/convex_gap_new.png}
%     \includegraphics[width=\linewidth]{imgs/convexity_gap_average_2x2.png}
%     \caption{Average convexity gap and exponential average convexity gap w.r.t. $\bx_{t-1}$. In most cases, the gaps are negative, indicating local convexity along the training trajectory.}
%     \label{fig:avggap}
%     \vspace{-1em}
% \end{figure}

Surprisingly, among all the tasks in our experiments, the average convexity gap along the optimization trajectories is consistently negative, except for the Bert pre-training task (\cref{fig:avggap}). This implies that the objective is locally convex in average along the training trajectory of SGDM and AdamW, despite its global non-convexity. This gives a hope that some convex optimization analysis could be still applied in practice, and we confirm this hypothesis in the next subsection.

Similarly, \citet{xing2018walk} observed that the loss interpolation $F(\alpha\bx_t+(1-\alpha)\bx_{t+1})$ for CIFAR-10 models trained by SGD is locally convex in $t$. We confirm this observation on different optimizers and a wider range of tasks.

As a remark, our results do not imply that the objective is locally convex along any trajectory---in fact, there must be a non-convex trajectory due to global non-convexity. Instead, our results provide a possible explanation for SGDM and AdamW's effectiveness, that is these optimizers can navigate the loss landscapes by following a good path that is ``sufficiently'' locally convex. 

% Further, as illustrated in Figure \ref{fig:avggap}, the dataset plays a significant role in shaping the loss landscape. Despite using the same optimizer settings and the ResNet-18 architecture, the loss landscapes for the ImageNet and CIFAR-10 datasets show markedly different levels of convexity. 
% \cref{fig:avggap} also shows that the average convexity gap across most experiments (except for Imagenet) indicates convex behavior along the optimization path. 
% This observation supports the applicability of convex optimization analysis, either with high probability or in expectation, for certain practical machine learning settings.

\subsection{Can Convexity-based Analysis Explain Optimization Success?}
\label{subsec:global-convex}

\begin{figure}[h]
    \centering
    \includegraphics[width=0.5\linewidth]{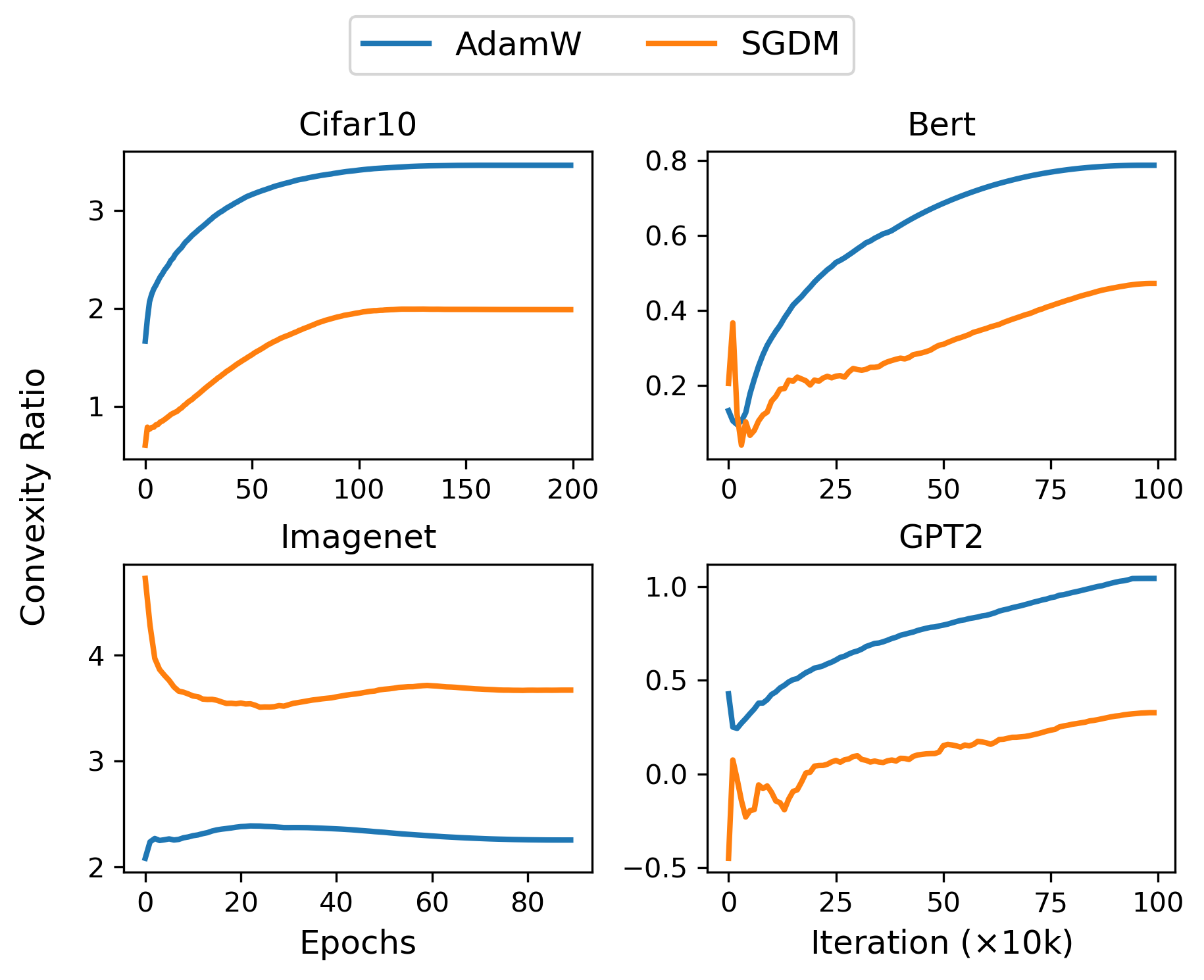}
    \caption{Convexity ratios of deep learning benchmarks where $\bx^\star$ is the final iterate from the same training run. 
    A convexity ratio greater than 1 indicates a convex function. Ratios between 0 and 1 suggest "weak quasi-convexity". Ratios less than 0 denote strong non-convexity. 
    See a complementary result where $\bx^\star$ is from a different training run in Appendix \ref{app:convex-ratio}.
    }
    \label{fig:convexityRatio}
\end{figure}

% Though the results in Section \ref{subsec:localconvex} suggest convexity along the optimization path often occurs, we might care more about global convexity, as this is useful to prove global convergence guarantees. Moreover, 
% While the convexity gap can be used to falsify convexity or give intuition about the local properties of the loss landscape, such ``local'' properties are less common in optimization analyses. So, 
Previous results show that the objective is often convex in average along training trajectory, thus suggesting a possibility that standard analyses might still hold in practice. In this section, we formalize this idea of being ``sufficiently'' convex with a new measure called \textit{convexity ratio}. At a high level, convexity ratio more directly measures the degree to which standard convex optimization analyses hold. Formally, we define:
\begin{align}
    \convRatio_T = \frac{\sum_{t=1}^T \langle \nabla F(\bx_t), \bx_t - \bx^\star\rangle}{\sum_{t=1}^TF(\bx_t) - F(\bx^\star)}.
    \label{eq:conv_ratio}
\end{align}
In practice, it is infeasible to compute this measure, so we approximate it as follows. First, we approximate the actual loss $F$ with a mini-batch loss with a large batch size. Second, since computing a large batch loss and its gradient is costly, we only compute the convexity ratio once every few iterations. See more details in Appendix \ref{subsec:expConfig}. 
Moreover, to better understand the landscape, we compute the convexity ratio w.r.t two different comparator $\bx^\star$: one is the final iterate from the same training run (Figure \ref{fig:convexityRatio}) and the other is from a training run with a different random seed (Figure \ref{fig:convex-ratio-diff}). 

When $F$ is convex, we should expect the convexity ratio to be larger than 1, which is implied by the following convexity inequality:
\begin{align}
    \textstyle
    \sum_{t=1}^TF(\bx_t) - F(\bx^\star) \le \sum_{t=1}^T \langle \nabla F(\bx_t), \bx_t - \bx^\star\rangle. 
    \label{eq:convexityBound}
\end{align}
\cref{eq:convexityBound} is the essential ingredient in convex optimization analyses.
% and is a cornerstone of the field of online convex optimization \citep{shalev2012online}, in which the left-hand side of this equation is called the ``regret'' \citep{hazan2022introduction}. 
In fact, many analyses directly provide an upper bound of the RHS, thus proving the convergence of an optimizer \citep{duchi10adagrad, mcmahan2010adaptive, zinkevich2003online, reddi2018convergence, hazan2007adaptive, hazan2006logarithmic}---such logic can be summarized by \cref{eqn:paradigm}. For example, a typical analysis of SGD \citep{zinkevich2003online} shows that
the RHS is bounded by $O(\sqrt{T})$,
% \begin{align*}
%     % \textstyle
%     \E\left[\sum_{t=1}^T \langle \nabla F(\bx_t), \bx_t - \bx^\star\rangle \right] \le O(\sqrt{T})
% \end{align*}
from which one can then conclude that $F(\overline\bx_T)-F(\bx^*)\le \frac{1}{T}\sum_{t=1}^TF(\bx_t) - F(\bx^\star)\le O(1/\sqrt{T})$, where $\overline \bx_T$ is the average of $\bx_1,\ldots,\bx_T$. 
% that is, the loss values of the iterates are ``on average'' approaching the loss of $F(\bx^\star)$. 
% This holds for all possible values of $\bx^\star$, even though we will only evaluate it for one particular point.

% and then rely on the fact that convexity implies that $\convRatio\ge 1$ to establish global convergence \cite{duchi10adagrad, zinkevich2003online, reddi2018convergence}.
% Then, to minimize the regret, it is enough to minimize the right-hand side, which is nothing
% else than the instantaneous linear regret on the linear function $\langle \nabla F(\bx_t), \cdot\rangle $ \citep{mcmahan2010adaptive,orabona2019modern}.
Even if $F$ is non-convex and \cref{eq:convexityBound} does not hold, it is still possible to derive global convergence so long as the convexity ratio is lower bounded by some $0< K <1$. Such condition is implied by ``weak quasi-convexity'' studied by \cite{orabona2017training}, and is formally stated as:
\begin{align}
    \textstyle
    \sum_{t=1}^TF(\bx_t) - F(\bx^\star) \le \frac{1}{K} \cdot \sum_{t=1}^T \langle \nabla F(\bx_t), \bx_t - \bx^\star\rangle. \label{eq:weaklyconvex}
\end{align}

Since common analyses usually bound the RHS of \cref{eq:convexityBound}, replacing convexity by weak quasi-convexity only degrades the convergence bound by a factor of $1/K$. Moreover, such analyses still ensure convergence as long as $K \ge \Omega(1/\sqrt{T})$. 

% Single Column
% \begin{wrapfigure}{r}{0.5\textwidth}
%     \centering
%     \vspace{-1em}
%     % \includegraphics[width=\linewidth]{imgs/ratio.png}
%     \includegraphics[width=\linewidth]{imgs/convexity_ratio_same_stationary_2x2.png}
%     \caption{Convexity ratios of deep learning benchmarks where $\bx^\star$ is the final iterate from the same training run. 
%     A convexity ratio greater than 1 indicates a convex function. Ratios between 0 and 1 suggest "weak quasi-convexity". Ratios less than 0 denote strong non-convexity. 
%     See a complementary result where $\bx^\star$ is from a different training run in Appendix \ref{app:convex-ratio}.
%     }
%     \label{fig:convexityRatio}
%     \vspace{-1em}
% \end{wrapfigure}

% Single Column (no wraparound)

% Double Column
% \begin{figure}
%     \centering
%     % \includegraphics[width=\linewidth]{imgs/ratio.png}
%     \includegraphics[width=\linewidth]{imgs/convexity_ratio_same_stationary_2x2.png}
%     \caption{Convexity ratios of deep learning benchmarks where $\bx^\star$ is the final iterate from the same training run. A convexity ratio greater than 1 indicates a convex function. Ratios between 0 and 1 suggest "weak quasi-convexity". Ratios less than 0 denote strong non-convexity. See a complementary result where $\bx^\star$ is from a different training run in Appendix \ref{app:convex-ratio}.}
%     \label{fig:convexityRatio}
%     \vspace{-1em}
% \end{figure}

Notably, we find that either convexity (Eq. \ref{eq:convexityBound}) or weak quasi-convexity (Eq. \ref{eq:weaklyconvex}) often hold in practice (\cref{fig:convexityRatio}). In the Imagenet experiments, both convexity ratios indicate convexity. For the CIFAR-10 experiments, AdamW's convexity ratio suggests the optimization trajectory remains globally convex relative to the stationary point. While SGDM shows slight non-convexity initially, its convexity ratio consistently exceeds 0.5, allowing for the application of classical convex analysis arguments. The same phenomenon applies to BERT and GPT2 experiments. These results suggest that the standard convex optimization analyses can be still applied to explain the empirical success of these optimizers, despite the globally non-convexity.

Furthermore, we observe that larger convexity ratio does not always correlate to better performance, e.g., AdamW has larger convexity ratio than SGDM in Cifar10 tasks but has lower test accuracy. On the other hand, the training performance gets worse as the convexity ratio becomes relatively small or even converges to 0 (e.g., SGDM on BERT and GPT2). Such observation agrees with the results from \cref{eq:weaklyconvex}. Interestingly, when we compute the convexity ratio w.r.t the final iterate of a different run, the result suggests the opposite (non-convexity). This suggests that the classical convexity analysis can help explain the success of practical optimization tasks but still has its limitations. Please see Appendix \ref{app:convex-ratio} for further discussion.
% However, that does not always seem to be a desirable property. 
% In Image Classification tasks, AdamW is worse than SGDM despite being able to find a more "globally convex" path.

% \begin{figure}[H]
% \centering
%  \begin{subfigure}
%      \centering
%      \includegraphics[width=0.24\textwidth]{imgs/ratio_cifar_same.png}
%   \end{subfigure}
%   \begin{subfigure}
%      \centering
%      \includegraphics[width=0.24\textwidth]{imgs/ratio_imagenet_same.png}
%   \end{subfigure}
%      \begin{subfigure}
%      \centering
%      \includegraphics[width=0.24\textwidth]{imgs/ratio_Bert_same.png}
%   \end{subfigure}
%    \begin{subfigure}
%      \centering
%      \includegraphics[width=0.24\textwidth]{imgs/ratio_gpt_same.png}
%   \end{subfigure}
% \caption{Convexity ratio
%     }
%   \label{fig:convexityRatio}
% \end{figure}

\section{Measuring Smoothness}
\label{sec:smoothness}

Smoothness assumptions plays a pivotal role in optimization theory. In convex optimization, smoothness can help accelerate the training process and achieve superlinear convergence rate if the loss is strictly convex or strongly convex \citep{nesterov2018lectures}. In non-convex optimization, smoothness is the key ingredient that makes many convergence analyses possible \citep{ghadimi2013stochastic,allen2016variance,jain2017non,reddi2019convergence}. Although smoothness is assumed for the majority of non-convex optimization results, it is unclear how well these smoothness conditions are satisfied in practice.

In fact, from a purely theoretical point of view, it may seem unlikely that the objective could be truly smooth: common activation functions such as the ReLU, and common layers such as MaxPools are not globally differentiable and so cannot possibly be smooth. However, one might hope that such issues are essentially pathological problems that do not affect practice - the loss could be ``smoothed out'' by averaging over a continuous data distribution for example. In this section, we attempt to measure smoothness along the real optimization trajectory in an efficient way analogous to our investigation of convexity in Section~\ref{sec:convex}.

% Single Column
% \begin{wrapfigure}{r}{0.5\textwidth}
%     \centering
%     \vspace{-1em}
%     % \includegraphics[width=\linewidth]{imgs/smooth_new.png}
%     \includegraphics[width=\linewidth]{imgs/smoothness_varying_lr_2x2.png}
%     \caption{Smoothness measures w.r.t. $\bx_{t-1}$. Experiments use optimal learning rate scheduler; see a complementary result that uses constant learning rates in Appendix \ref{app:smoothness}.}
%     % (Top) are the experiments with optimal learning rate scheduler, and (bottom) are the experiments with constant learning rate. Details of experiment setup can be found in Appendix \ref{subsec:expConfig}.}
%     \label{fig:smooth}
%     \vspace{-1em}
% \end{wrapfigure}

% Single Column (no wraparound)
\begin{figure}[h]
    \centering
    \includegraphics[width=0.5\linewidth]{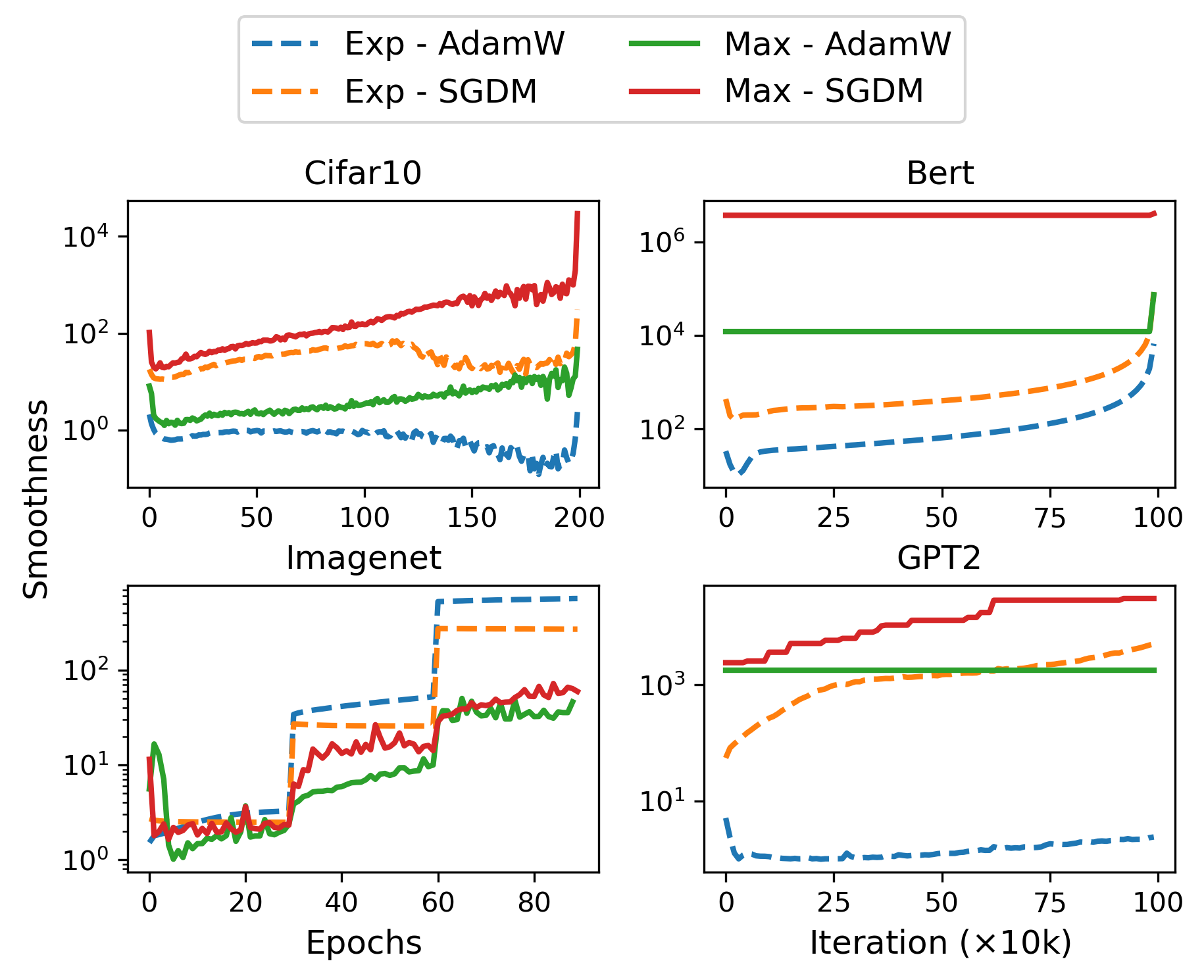}
    \caption{Smoothness measures w.r.t. $\bx_{t-1}$. Experiments use optimal learning rate scheduler; see a complementary result that uses constant learning rates in Appendix \ref{app:smoothness}.}
    % (Top) are the experiments with optimal learning rate scheduler, and (bottom) are the experiments with constant learning rate. Details of experiment setup can be found in Appendix \ref{subsec:expConfig}.}
    \label{fig:smooth}
\end{figure}

We measure the exponential average and the maximum smoothness defined in \cref{eq:avg_exp-smooth} since they capture the smoothness level of local and global loss landscape respectively. 
As in \cref{fig:smooth} (top), in all experiments, the smoothness constants appear to be upper-bounded. However, in many cases these constants are quite large ($10^3$ to $10^6$), making it hard to consider the loss landscapes in these experiments to be smooth in practice. 
Furthermore, we note that smoothness correlates with changes in the learning rate scheduler. For example, as the learning rate approaches zero at the end of training, the smoothness value increases, as observed in Cifar10 with cosine decay and BERT with linear decay. Similarly, for Imagenet, where we used a piecewise linear scheduler, smoothness increases whenever the learning rate decreases. This observation suggests that smaller learning rates tend to result in larger smoothness values. 
% Hence, the results in \cref{fig:smooth} (top) are significantly influenced by both the warmup and decay phases of the learning rate scheduler. % seems a bit vague.

% To better understand the loss landscapes, we reran all experiments with a constant learning rate (\cref{fig:smooth} bottom).
% With constant learning rates, the loss landscape appeared smoother and more stable. Both the max and exponential average smoothness followed a similar pattern: a rapid drop initially (except for SGDM on ImageNet), followed by a consistent rise until reaching a boundary, then stabilizing. Adam typically achieved smaller (i.e., smoother) measures with a learning rate scheduler, while SGD found smaller measures with a constant rate. We conjecture that this phenomenon suggests that SGD's optimization path is more sensitive to changes in the learning rate, while Adam remains robust across different learning rate settings.

\subsection{Smoothness Measures as Proxies for Sharpness}
\label{subsec:sharpness}

The smoothness measure in \cref{fig:smooth} exhibit similar behaviors across different experiments, which resembles the edge-of-stability phenomenon observed by \citep{cohen2020gradient, cohen2022adaptive} in GD and full-batch Adam for smaller tasks. Specifically, \cite{cohen2020gradient} defined the ``sharpness'' as the operator norm of the Hessian $\nabla^2 F(\bx_t)$, and they observed that when training with GD on CIFAR-10, the sharpness increases up to some value inversely proportional to the learning rate, and then stabilizes.

Our measurements track different quantities than the sharpness, but are faster to compute. Thus, these observations pose an interesting question: \textit{Can our new metrics, $\maxSmooth$ and $\expSmooth$, be used as proxies for the sharpness}? If this is true, our approach could substantially expedite the evaluation of sharpness. Our method also makes evaluating the sharpness of much larger models possible (for which computing Hessian information is prohibitively expensive).

% Single Column
% \begin{wrapfigure}{r}{0.5\textwidth}
%     \centering
%     \vspace{-1em}
%     \includegraphics[width=\linewidth]{imgs/sharp_convex_new.png}
%     \caption{Sharpness v.s. Smoothness Measures.}
%     \label{fig:sharpness}
%     \vspace{-1em} % hack to fix "margin overflow" issue
% \end{wrapfigure}

% Single Column
\begin{figure}[h]
    \centering
    \vspace{-1em}
    \includegraphics[width=0.5\linewidth]{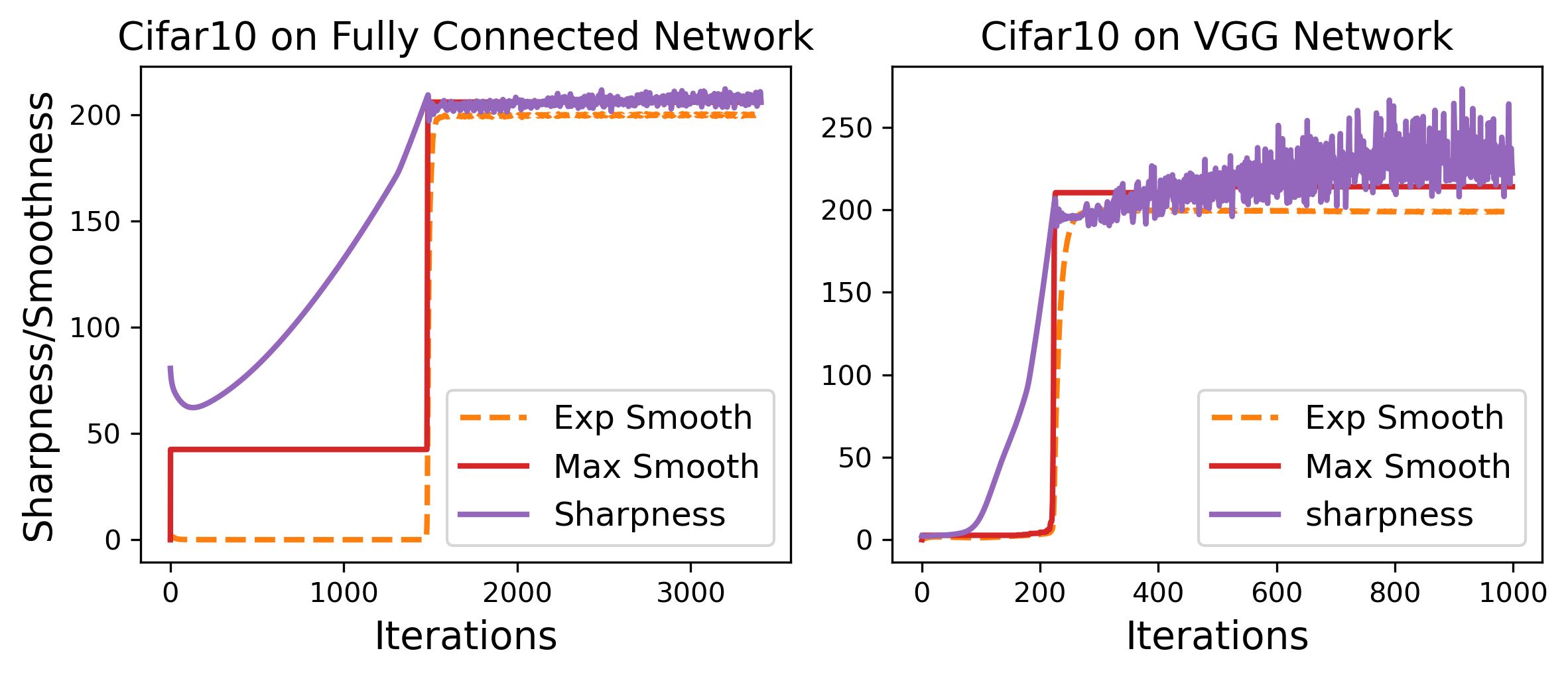}
    \caption{Sharpness v.s. Smoothness Measures.}
    \label{fig:sharpness}
    \vspace{-1em} % hack to fix "margin overflow" issue
\end{figure}

% Double Column
% \begin{figure}[ht]
%     \centering
%     \includegraphics[width=\linewidth]{imgs/sharp_convex_new.png}
%     \caption{Sharpness v.s. Smoothness Measures.}
%     \label{fig:sharpness}
%     \vspace{-1em} % hack to fix "margin overflow" issue
% \end{figure}

As discussed above, we notice that a smaller learning rate results in a larger smoothness value. We can potentially explain this using the edge-of-stability phenomenon. \citep{cohen2020gradient, cohen2022adaptive} observe that the sharpness is oscillating at the value $c/\eta$ for some constant $c>0$ and $\eta$ is the learning rate at the edge of stability. Thus, when the learning rate scheduler is applied, any time the learning drops, this boundary increases and causes the smoothness/sharpness level to increase. This phenomenon is also observed in \citep{cohen2022adaptive}. To verify our conjecture, we replicate the experiments in \citep{cohen2020gradient} where we train Cifar10 on a simple linear network with tanh activation and on a VGG-11 network \citep{simonyan2014very} in Fig.\ref{fig:sharpness}.

% \begin{figure}
% \centering
%  \begin{subfigure}
%      \centering
%      \includegraphics[width=0.48\textwidth]{imgs/sharp_tanh.png}
%   \end{subfigure}
%   \begin{subfigure}
%      \centering
%      \includegraphics[width=0.48\textwidth]{imgs/sharp_vgg.png}
%   \end{subfigure}
%     \caption{Sharpness (maximum eigenvalue of the training loss Hessian Matrix) vs Smoothness}
%     \label{fig:sharpness}
% \end{figure}
Our new smooth metrics track the actual sharpness value very closely (Fig.\ref{fig:sharpness}). One possible justification for this is when we measure $\frac{\|\nabla f(\bx_t,z) - \nabla f(\bx_{t-1},z)\|}{\|\bx_t - \bx_{t-1}\|}$, we are effectively estimating how quickly the gradient of the function changes, which is bounded by the Hessian's spectral norm in smooth functions. A higher value indicates a steeper change in the gradient, implying a larger maximum eigenvalue of the Hessian matrix, hence a higher "sharpness". Thus, this metric and the sharpness are inherently related to characterizing the function’s smoothness and curvature. 

% We note that our results in Fig.\ref{fig:sharpness} still have some inconsistency with previous works. For instance, \citep{cohen2022adaptive} suggests that Adam finds sharper minima than SGD, which we did not observe in our Cifar10 experiment. However, due to the different experimental settings (different network, fullbatch vs small minibatch), it is difficult to conclude whether these inconsistencies are due to the metrics or the experiments settings. Therefore, a more extensive investigation is needed to determine if our smoothness metrics can serve as proxies for sharpness. We will leave this for future research.

\subsection{Can Smoothness-based Analysis Explain Optimization Success?}
\label{subsec:update-corr}

% From the above results, it seems reasonable to assume that our loss landscape is smooth, albeit with a fairly large smoothness constant. 
% (especially when a decaying learning rate scheduler is employed). 
The smoothness measurements discussed above are not actually the best criterion for judging the applicability of smooth non-convex optimization analysis. This is because they only capture gradient behavior rather than linking gradients to function values. In typical smoothness-based analysis, one encounters the quantity $\langle \nabla f(\bx_{t+1},z_{t+1}), \bx_{t+1}-\bx_{t}\rangle$. In almost all analyses of non-convex optimization algorithms, this quantity usually plays the role of the ``algebraic expression'' in  (\ref{eqn:paradigm})  \citep{khaled2020better,li2024convex,NEURIPS2018_90365351,doi:10.1137/17M1114296, li2019convergence, faw2022power, reddi2019convergence}. To illustrate, consider an optimizer with update $\bx_{t+1} = \bx_t + \Delta_t$, and assume that $F$ is $L$-smooth, $\E[\Delta_t] = -\eta\nabla F(\bx_t)$ and $\E[\|\Delta_t\|^2]\le \eta^2G^2$. Then:
\begin{align}
    \E[F(\bx_{t+1}) - F(\bx_t)] 
    &\textstyle \le \E[ \langle \nabla F(\bx_t), \bx_{t+1} - \bx_t\rangle + \frac{L}{2}\|\bx_{t+1} - \bx_t\|^2 ] \notag\\
    % \intertext{Assuming $\E[\Delta_t] = \nabla F(\bx_t)$ and $\Delta_t$ has variance at most $\sigma^2$ for all $t$, then:}
    % F(\bx_{t+1}) 
    % &\textstyle \le -\eta\E\left[\|\nabla F(\bx_t)\|^2\right] + \frac{L\eta^2}{2}\E\left[\|\nabla F(\bx_t)\|^2\right] + \frac{L\eta^2\sigma^2}{2} \notag\\
    % \intertext{Let $\eta \le \frac{1}{H}$:}
    % F(\bx_{t+1}) 
    &\textstyle \le -\eta\E\left[\|\nabla F(\bx_t)\|^2\right] + \frac{L\eta^2G^2}{2}. 
    \label{eq:descentLemma}
\end{align}
Typical analyses show that $-\eta\E[\|\nabla F(\bx_t)\|^2]$ dominates $\frac{L\eta^2G^2}{2}$ so that $F(\bx_t)$ decreases over time. Intuitively, this holds if we make $\eta$ sufficiently small because the negative term is linear in $\eta$ while the positive term is quadratic in $\eta$. Note that this high-level idea is used even for analyses based on less classical smoothness assumptions such as (L0,L1) smoothness \citep{zhang2019gradient}.

To check whether this analysis technique can explain the success of practical optimizers, we would like to measure the inner-product $\langle \nabla F(\bx_t), \bx_{t+1}-\bx_t\rangle$ and see if it is negative. This would directly capture the optimization analysis because in the typical analysis, all of the provable decrease in the function value is caused by negative inner-products.

Unfortunately, this inner-product is difficult to estimate empirically because we do not know $\nabla F(\bx_t)$. 
% One might consider instead estimating it using $\langle \nabla f(\bx_t, z_t), \bx_{t+1} - \bx_t \rangle$. However, this approach is flawed because $\bx_{t+1} - \bx_t$ is not independent of $z_t$, giving the correlation a negative bias. 
Instead, we measure a quantity that we call the \textit{update correlation}, which is defined as 
\begin{equation}
    \gdx_t \coloneqq \langle \nabla f(\bx_{t+1}, z_{t+1}), \bx_{t+1}-\bx_t\rangle. 
    \label{eq:update_corr}
\end{equation}
Since $\bx_{t+1}-\bx_t$ is independent of $z_{t+1}$, the update correlation is an unbiased estimator of $\langle \nabla F(\bx_{t+1}), \bx_{t+1}-\bx_t\rangle$. Moreover, it turns out that update correlation still captures the same notion of ``function'' progress measured by typical analysis. Consider the update $\bx_{t+1} = \bx_t + \Delta_t$ and assume $F$ is $L$-smooth. Then,
% \begin{align}
%     F(\bx_{t+1}) - F(\bx_t)
%     &\textstyle \ge \langle \nabla F(\bx_t), \bx_{t+1} - \bx_t\rangle - \frac{L}{2}\|\bx_{t+1} - \bx_t\|^2 \notag\\
%     &\textstyle = \langle \nabla F(\bx_{t+1}), \bx_{t+1}-\bx_t \rangle +  \langle \nabla F(\bx_t) - \nabla F(\bx_{t+1}), \Delta_t \rangle - \frac{L}{2}\|\Delta_t\|^2 \notag\\
%     &\textstyle \ge \langle \nabla F(\bx_{t+1}), \bx_{t+1}-\bx_t\rangle - \frac{3L}{2}\|\bx_{t+1}-\bx_t\|^2.
%     \label{eq:positiveDescent}
% \end{align}
\begin{align}
    & F(\bx_{t+1}) - F(\bx_t) 
    \textstyle \ge \langle \nabla F(\bx_{t+1}), \bx_{t+1} - \bx_t\rangle - \frac{L}{2}\|\bx_{t+1} - \bx_t\|^2, \label{eq:positiveDescent}\\
    & F(\bx_{t+1}) - F(\bx_t) 
    \textstyle \le \langle \nabla F(\bx_{t+1}), \bx_{t+1} - \bx_t\rangle + \frac{L}{2}\|\bx_{t+1} - \bx_t\|^2. \label{eq:positiveDescentUB}
\end{align}
Thus, if $\langle \nabla F(\bx_{t+1}), \bx_{t+1}-\bx_t\rangle$ is negative, for small enough learning rates $\eta$ the global loss decreases and the optimizer is consistently making progress. On the other hand, a positive update correlation $\langle \nabla F(\bx_{t+1}), \bx_{t+1}-\bx_t\rangle$ appears to be disastrous since this analysis would suggest that the loss should increase. In particular, we are not aware of any analysis based upon negative values of $\langle \nabla F(\bx_t), \bx_{t+1} -\bx_t\rangle$ that does not also predict negative values for the update correlation.
% can be disastrous since $\frac{L}{2}\|\bx_{t+1}-\bx_t\|^2$ is only $O(\eta^2)$. Thus, if the update correlation is substantial enough to outweigh this negative term, the loss might not decrease at all and the smooth non-convex analysis falls apart. 
Therefore, if the standard analysis of smooth non-convex optimization can explain optimization success in deep learning, then in every experiment we should expect that the update correlation $\langle \nabla f(\bx_{t+1},z_{t+1}), \bx_{t+1}-\bx_{t}\rangle$ should be negative on average. 

% Single Column
% \begin{wrapfigure}{r}{0.5\textwidth}
%     \centering
%     \vspace{-1em}
%     \includegraphics[width=\linewidth]{imgs/update_corr_convex_new.png}
%     \caption{Update correlation of GD on the squared loss (left) and logistic regression on OpenML datasets (right). The blurred lines are the actual update correlations, and the thick lines are the average.}
%     % the average update correlations.}
%     \label{fig:linear_convex}
%     \vspace{-1em}
% \end{wrapfigure}

% Single Column
\begin{figure}[h]
    \centering
    \includegraphics[width=0.5\linewidth]{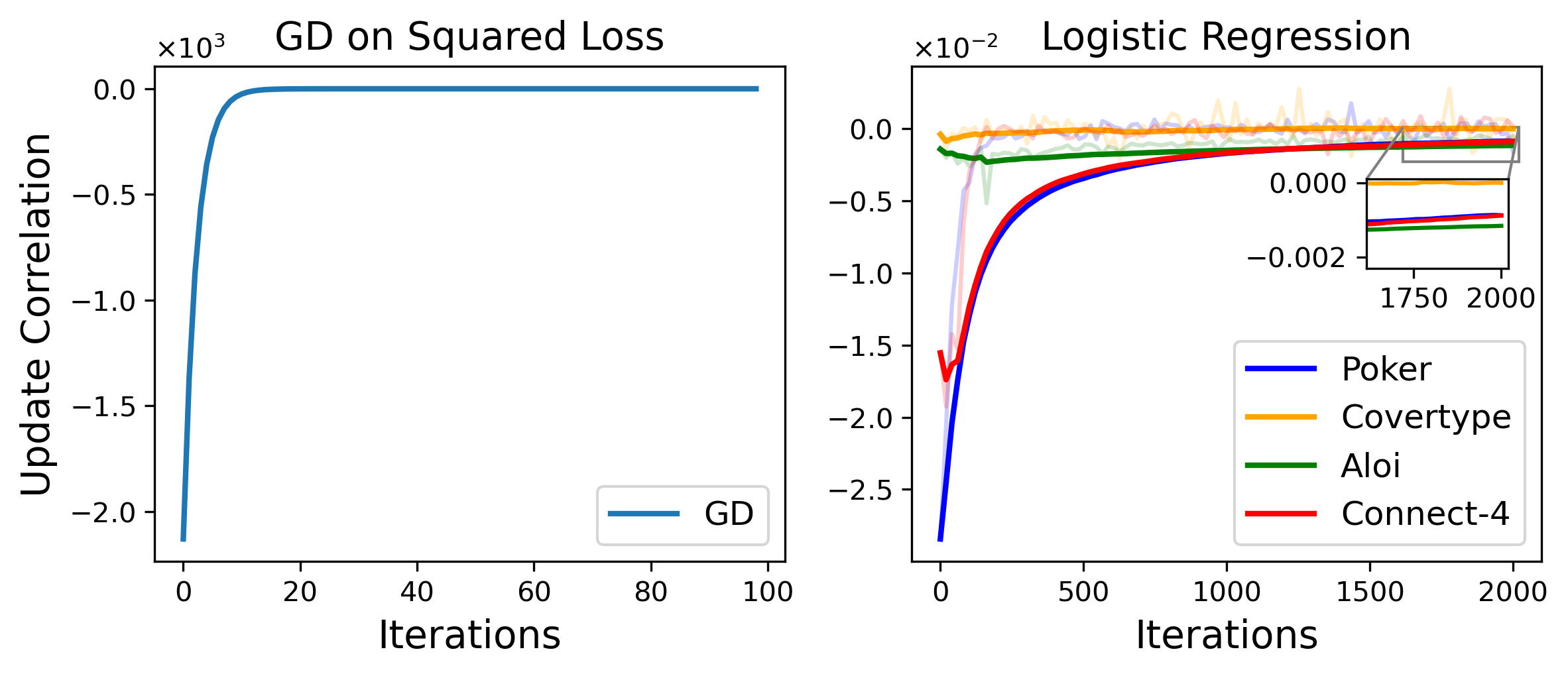}
    \caption{Update correlation of GD on the squared loss (left) and logistic regression on OpenML datasets (right). The blurred lines are the actual update correlations, and the thick lines are the average.}
    % the average update correlations.}
    \label{fig:linear_convex}
\end{figure}

% Double Column
% \begin{figure}[ht]
%     \centering
%     \includegraphics[width=\linewidth]{imgs/update_corr_convex_new.png}
%     \caption{Update correlation of GD on the squared loss (left) and logistic regression on OpenML datasets (right). The blurred lines are the actual update correlations, and the thick lines are the average.}
%     % the average update correlations.}
%     \label{fig:linear_convex}
%     \vspace{-1em}
% \end{figure}

\begin{figure}[h]
    \centering
    \includegraphics[width=0.75\linewidth]{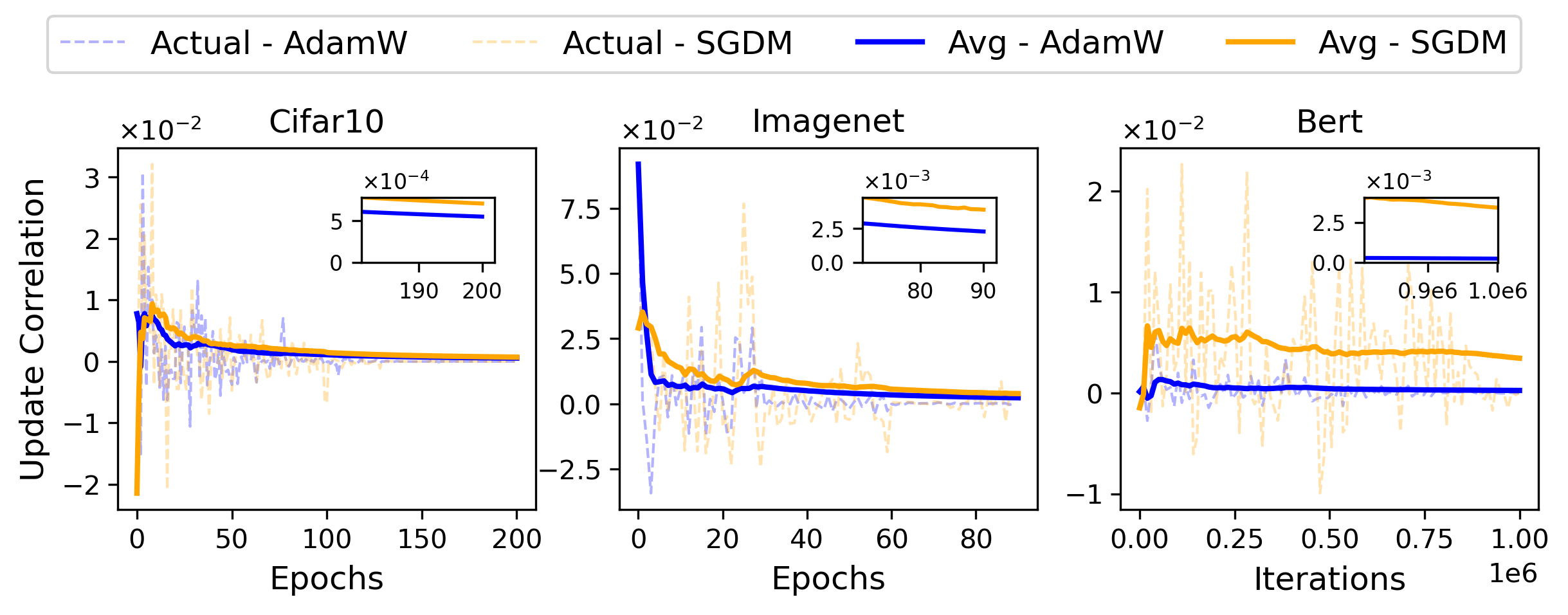}
    \caption{Update correlation on deep learning tasks.}
    \label{fig:linearcor}
\end{figure}

First, we check if this is the case for simple convex experiments (\cref{fig:linear_convex}). In all of these experiments, the update correlations are negative on average, which agrees with our intuition.
% Initially, when the optimizer quickly progresses toward the minimum, the linear correlation is highly negative. As training progresses and the optimizer begins to oscillate around the minimum, this value converges towards zero.
However, surprisingly, the update correlation is positive on average in almost every other Deep Learning experiment (\cref{fig:linearcor}). This is a fascinating phenomenon because it indicates that the optimizer changes direction very often, and
yet it still effectively minimizes the loss. This suggests that the classic smooth non-convex analysis that relies on the descent lemma is problematic in practice. The only case of negative correlations is GPT2 on the Pile dataset, but they turn positive when the dataset is shuffled or replaced with the C4 dataset. It would be interesting to find out exactly the cause of this behavior. 

% Single Column
% \begin{wrapfigure}{r}{0.5\textwidth}
%     \centering
%     \vspace{-1em}
%     % \includegraphics[width = \linewidth]{imgs/update_corr.png}
%     \includegraphics[width = \linewidth]{imgs/update_correlation_1x3.png}
%     \caption{Update correlation on deep learning tasks.}
%     \label{fig:linearcor}
%     \vspace{-2em}
% \end{wrapfigure}

% Single Column

% Double Column
% \begin{figure}[ht]
%     \centering
%     % \includegraphics[width = \linewidth]{imgs/update_corr.png}
%     \includegraphics[width = \linewidth]{imgs/update_correlation_1x3.png}
%     \caption{Update correlation on various Deep Learning tasks.}
%     \label{fig:linearcor}
%     \vspace{-1em}
% \end{figure}

% \subsection{Lower bound on (5)}

% First, by lower bound of smoothness, we have
% \begin{align*}
%     F(\bx_{t+1}) 
%     &\ge F(\bx_t) + \langle g_t, \bx_{t+1} - \bx_t\rangle - \frac{L}{2}\|\bx_{t+1} - \bx_t\|^2 \notag\\
%     &= F(\bx_t) + \langle g_{t+1}, \bx_{t+1} - \bx_t\rangle +  \langle g_t - g_{t+1}, \bx_{t+1} - \bx_t\rangle - \frac{L}{2}\|\bx_{t+1}-\bx_t\|^2.
%     \label{eq:positiveDescent}
% \end{align*}
% Next, again by smoothness,
% \begin{align*}
%     \langle g_t - g_{t+1}, \bx_{t+1} - \bx_t\rangle - \frac{L}{2}\|\bx_{t+1}-\bx_t\|^2
%     &\ge -\|g_t-g_{t+1}\|\|\bx_{t+1}-\bx_t\| - \frac{L}{2}\|\bx_{t+1}-\bx_t\|^2 \\
%     &\ge -\frac{3L}{2}\|\bx_{t+1}-\bx_t\|^2.
% \end{align*}
% Combining both, we have
% \begin{equation*}
%     F(\bx_{t+1}) - F(\bx_t) \ge \langle g_{t+1}, \bx_{t+1}-\bx_t\rangle - \frac{3L}{2}\|\bx_{t+1}-\bx_t\|^2.
% \end{equation*}

% \subsubsection{Conditioning}\label{sec:conditioning}
The observation that $\nabla F(\bx_{t+1})$ is positively correlated with $\bx_{t+1}-\bx_t$ suggests that the objective may be ``poorly conditioned'',  so that the optimizer is bouncing back-and-forth along the walls of a narrow ravine in the optimization landscape. Previous empirical studies have also suggested similar dynamics \citep{rosenfeld2023outliers}. The classical mitigations for poorly conditioned objectives in the \emph{deterministic or convex} settings are preconditioning, including via second-order algorithms, as well as accelerated gradient descent (e.g. \cite{gupta2018shampoo, liu2023sophia,yao2021adahessian, nesterov2018lectures,dozat2016incorporating}). However, the advantages of such techniques are poorly understood in the stochastic setting (indeed, there is no advantage in the worst-case \citep{arjevani2020second}). Instead, most current analyses we are aware of in the  stochastic setting appear to rely on negative update correlations. 

% Non-smooth subsection
\subsection{Verification of Non-smooth Optimization}
\label{sec:non-smooth}

% There have been various theoretical works on relaxing the smoothness assumption. However, the absence of smoothness introduces several significant challenges. Firstly, as previously mentioned, smoothness ensures that subsequent gradients exhibit relatively small changes, provided the parameters do not move too much. In contrast, in the non-smooth scenario, gradients can change drastically, invalidating the standard descent lemma based on Taylor approximation. Secondly, prior research has demonstrated that, in the worst case, it is impossible to find first-order stationary points when the objective function is non-smooth \citep{kornowski2022oracle}. These challenges lead to two natural motivations: the need for a tractable notion of convergence criteria and the necessity for more advanced analytical techniques beyond the descent lemma based on Taylor approximation.
% In previous sections, we observed that some common assumptions or identities used in analysis, such as convexity, smoothness, or negative update correlation, might not hold in practice. In this section, we will discuss alternative frameworks that do not rely on these assumptions.
% To better understand the training properties, we turn to the theoretical literature on non-smooth, non-convex optimization, aiming to draw a connection between theory and practice.
% Below we present a brief summary of the theoretical works on non-smooth non-convex optimization. 
Recently there have been works that analyze optimization algorithms in the absence of smoothness assumptions. One direction proposes relaxed smoothness assumptions \citep{zhang2019gradient, liu2023sophia}. Another direction studies weakly convex objectives \citep{davis2019stochastic, mai2020convergence}, aiming to minimize a proxy of the objectives called the Moreau envelope \citep{moreau1965proximite}.
% Specifically, a function $F$ is $\rho$-weakly convex if $F(\bx)+\frac{\rho}{2}\|\bx\|^2$ is convex, and its Moreau envelope \citep{moreau1965proximite} with parameter $\lambda$ is defined as $F_\lambda(\bx) = \min_\by (F(\by) + \frac{1}{2\lambda}\|\by-\bx\|^2)$. 
% Intuitively, the Moreau envelope serves as a smoothing of the non-smooth objective. 
% Prior works usually focused on finding first-order $\epsilon$-stationary points of the Moreau envelope $F_\lambda$ \citep{davis2019stochastic, mai2020convergence}. Rather than requiring a negative inner-product $\langle \nabla F(\bx_t), \bx_{t+1}-\bx_t\rangle$, this style of analysis often requires a negative inner-product of the form $\langle \text{Prox}_{F}(\bx_t) - \bx_t , \bx_{t+1}-\bx_t\rangle$, where $\text{Prox}_F$ is the proximal operator $\text{Prox}_F(x) = \argmin_y F(y) + \lambda \|x-y\|^2$ for some appropriate $\lambda$. This identity is challenging to verify because estimating the proximal operator seemingly requires solving an optimization problem itself. We leave an empirical tractable verification of this identity as an important open problem.
There's also a direction that adopts the Goldstein stationary point \citep{goldstein1977optimization} as a convergence criterion that is tractable for non-smooth objectives \citep{zhang2020complexity, cutkosky2023optimal, ahn2024understanding, zhang2024random}. 

In this section, we verify the identities central to the analysis of the last line of works.
Specifically, \citet{cutkosky2023optimal} proposes an online-to-non-convex conversion (O2NC) technique, and its key idea the use of random scaling: 
suppose $s_t$ is sampled i.i.d. from $\mathrm{Exp}(1)$, then $\bx_{t+1} = \bx_t+s_t\Delta_t$ satisfies $\Ex_{s_t}[F(\bx_{t+1}) - F(\bx_t)] = \Ex_{s_t}[\langle \nabla F(\bx_{t+1}), \Delta_t \rangle]$.

% Single Column
% \begin{wrapfigure}{r}{0.5\textwidth}
%     \centering
%     \vspace{-1em}
%     \includegraphics[width=\linewidth]{imgs/gpt_pile_nonsmooth.png}
%     \caption{Cumulative sum of update correlation, update correlation with RS, and loss difference of GPT2 experiments. (Top) is SGDM and (bottom) is AdamW; (left) is update with RS and (right) without RS.}
%     \label{fig:gpt-o2nc}
%     \vspace{-1em}
% \end{wrapfigure}

% Single Column
\begin{figure}[h]
    \centering
    \vspace{-1em}
    \includegraphics[width=0.5\linewidth]{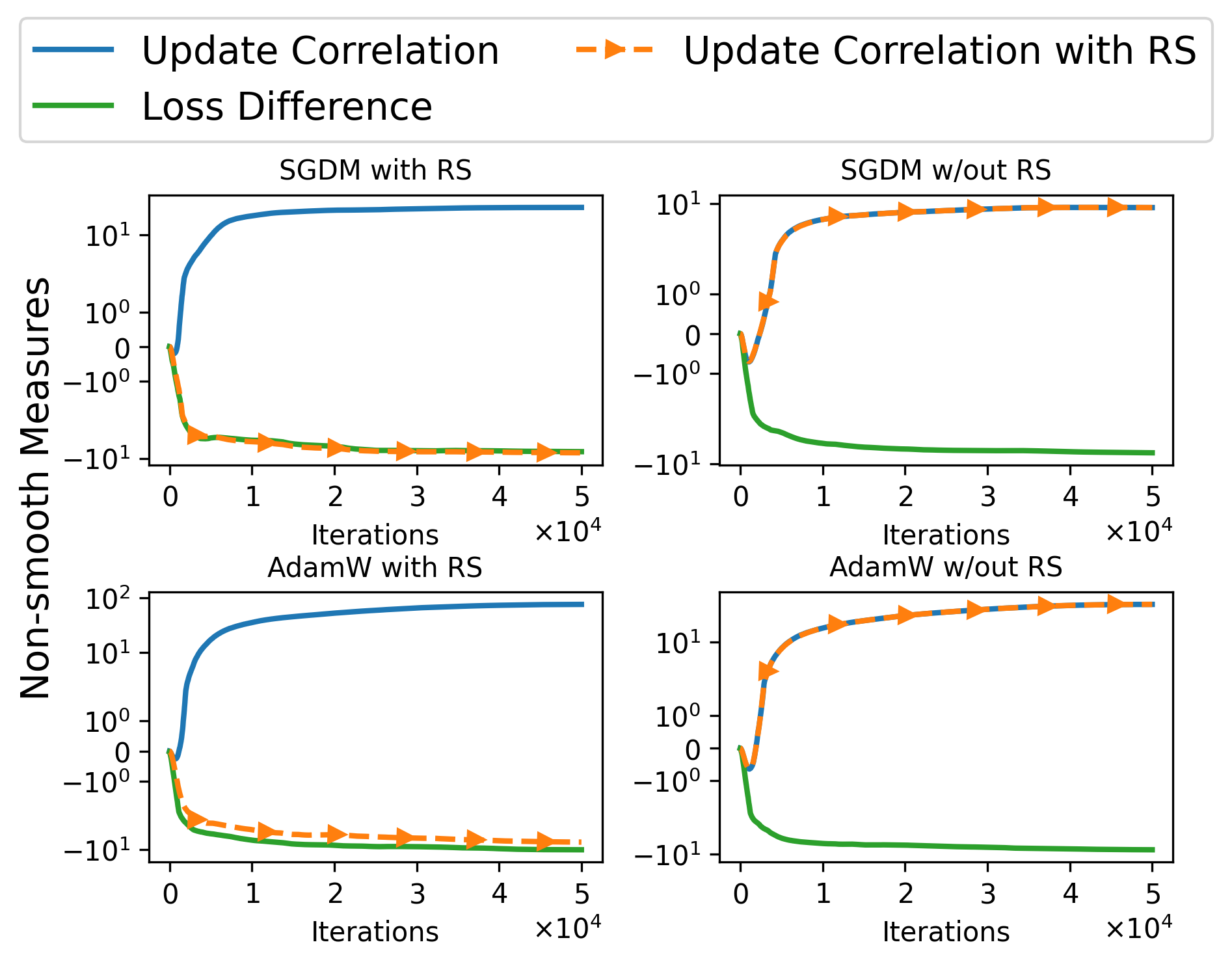}
    \caption{Cumulative sum of update correlation, update correlation with RS, and loss difference of GPT2 experiments. (Top) is SGDM and (bottom) is AdamW; (left) is update with RS and (right) without RS.}
    \label{fig:gpt-o2nc}
    \vspace{-1em}
\end{figure}

% Double Column
% \begin{figure}[ht]
%     \centering
%     \includegraphics[width=\linewidth]{imgs/gpt_pile_nonsmooth.png}
%     \caption{Cumulative sum of update correlation, update correlation with RS, and loss difference of GPT2 experiments. (Top) is SGDM and (bottom) is AdamW; (left) is update with RS and (right) is the benchmark without RS.}
%     \label{fig:gpt-o2nc}
%     % \vspace{-1em}
% \end{figure}

We refer to the update form $\bx_{t+1} = \bx_t + s_t\Delta_t$ where $s_t \sim \text{Exp}(1)$ i.i.d. as the \textit{update with random scaling (RS)}, and the update with $s_t\equiv 1$ as the \textit{update without RS}.
Unlike the lower bound in \eqref{eq:positiveDescent},the above equality suggests that $\langle \nabla F(\bx_{t+1}),\Delta_t\rangle$, which we referred to as \textit{update correlation with RS}, is an unbiased estimator of function progress $F(\bx_{t+1})-F(\bx_t)$ and a good indicator of the training progress: we should expect $F(\bx_t)$  to decrease as long as $\langle \nabla F(\bx_{t+1}),\Delta_t\rangle$ is negative in average. To verify if the theory holds in practice, we test SGDM and AdamW with random scaling updates and compare them to their counterparts without RS. Specifically, we measure the following three properties: update correlation, update correlation with random scaling, and instantaneous loss difference, where the first is defined in Eq. \ref{eq:update_corr} and the latter two are respectively defined as
\begin{align}
    &\gdelta_t = \langle \nabla f(\bx_t,z_t), \Delta_{t-1}\rangle, \notag\\ 
    &\dloss_t = f(\bx_t,z_t) - f_t(\bx_{t-1},z_t).
    \label{eq:update_corr_RS}
\end{align}
Note that if the update does \emph{not} have random scaling applied, then $\gdelta_t=\gdx_t$.
% Note that for update with random scaling, $\bx_t-\bx_{t-1} = s_{t-1}\Delta_{t-1}$ and thus $\gdx_t = s_{t-1} \times \gdelta_t$; for update without RS, $\bx_t-\bx_{t-1} = \Delta_{t-1}$ and $\gdx_t = \gdelta_t$. 

\cref{fig:gpt-o2nc} plots cumulative sum of these quantities. The sum of update correlation always increases, regardless of whether random scaling is employed. However, for optimizers with random scaling, the sum of $\gdelta_t$ decreases and closely aligns with the sum of loss difference. This supports the theory that $\gdelta_t$ is an unbiased estimator of loss difference, even for complicated LLM pre-training tasks where the objective is non-convex and non-smooth. 
% Also, it motivates a guideline for developing empirically effective optimizers: keeping $\langle \nabla f(\bx_t,z_t),\Delta_{t-1}\rangle$ as negative as possible while applying random scaling to the update.
% Our primary concern in practice is effectively reducing the loss value, and any optimizer with sufficiently negative update correlation with RS, once equipped with random scaling, should achi great function progress, 
\section{Conclusions}
In this paper, we address the critical question of whether modern analyses in stochastic optimization theory align with practice. To this end, we empirically measure key quantities that are commonly used in theory across a diverse range of machine learning benchmarks. Our results indicate that although the loss objective is globally non-convex, the convexity ratio is typically positive and thus convex optimization analyses may still be applied. On the other hand, the objective is non-smooth neither globally nor locally, and smoothness-based analyses often do not hold. We hope that our experiments results can contribute to a better understanding of what enables practical optimization, as well as motivate more rigorous empirical verification of optimization analyses in the future.

We admit our work has several limitations. First, we focus exclusively on evaluating existing optimization analyses rather than proposing new algorithms or assumptions. While exploring new assumptions that more closely reflect the practice is valuable, we believe that rigorously verifying which theoretical results hold, and which do not, in real training scenarios is equally important. Second, due to resource constraints, our experiments are limited to the two most widely used optimizers (SGDM and AdamW); extending our evaluation to other optimizers is left for future work.

% \textbf{Limitations:} 
% Most of our results are based on "local" quantities since we use past iterates as the reference point. This raises the question: is there a better way to efficiently approximate "global" quantities? Achieving this would provide a more comprehensive view of the loss landscape. Furthermore, while we propose some non-convex, non-smooth methods as alternatives, their robustness and practical applicability remain uncertain. Developing other frameworks would be an interesting avenue for future research. Additionally, some of our conjectures have not been fully investigated. For instance, the smoothness/sharpness comparison has only been conducted on a few simple image classification tasks. We leave this for future research.

%%%%%%%%%%%%%%%%%%%%%%%%%%%%%%%%%%%%%%%%%%%%%%%%%%%%%%%%%%%%

\bibliography{reference}
\bibliographystyle{apalike}

%%%%%%%%%%%%%%%%%%%%%%%%%%%%%%%%%%%%%%%%%%%%%%%%%%%%%%%%%%%%

\newpage
\appendix

% \appendix
\section{License}
\label{app:license}
Image Classification: Imagenet is distributed under the BSD 3-Clause License, Resnet is distributed under the Apache License, Cifar10 is distributed under the MIT License.

NLP: Bert/GPT2 (Hugging Face), C4 dataset are distributed under the Apache 2.0 License. The Pile dataset is distributed under the MIT License.

\section{Experiments Settings and Configurations}
\label{subsec:expConfig}
\textbf{GD on squared loss:} We run Gradient Descent on squared loss using synthetic datasets. We train the optimizer for 100 iterations with a learning rate equals to 0.1. We report the metrics computed every iteration.

\textbf{Logistic regression with OpenML datasets:} We run logistic regression with commonly used OpenML datasets such as Aloi (42396), Poker (1595), Connect-4 (1591), and Covertype (150). We run all experiments using a batch size equal to 64 and AdamW as the optimizer. We tune the learning rates using a grid search over the range $[1e-4, 1]$. We report the metrics computed at the end of every epoch.

\textbf{Training Cifar10 on Resnet18:} We train the benchmark dataset Cifar10 on Resnet18 using SGDM and Adamw. For SGDM, we use a learning rate $=0.1$ and for AdamW, we use a learning rate $=0.001$. Both optimizers are trained with batch size equal to 128, weight decay equal to $5e-4$, and cosine learning rate scheduler. In the experiments with constant learning rates, we use the same optimal configurations as the normal experiments but without the scheduler. We train both optimizers for 200 epochs and all tracking measures (convexity gap, max smoothness, etc,...) are reset for the new epoch (this is why we see the max smoothness quantity goes down at various points in Fig.\ref{fig:smooth}). We use full batch to compute the large batch loss ($F(x)$) and gradient $\nabla F(x)$. We report the metrics computed at the end of every epoch.

\textbf{Training Imagenet on Resnet18:} We train Imagenet on Resnet18 using SGDM with a learning rate equal to 0.1 and Imagenet with a learning rate equal to 0.001. The weight decay is $1e-4$ and we employ a learning rate scheduler that decays the learning rate by 10 every 30 epochs for both optimizers. These are the experiments configurations used in \citep{yao2020adahessian,tran2022better}. Similar to the Cifar10 experiments, we keep the same configurations except for the learning rate scheduler for the constant learning rates experiments. We also reset the tracking quantities every epoch. We use full batch to compute the large batch loss ($F(x)$) and gradient $\nabla F(x)$. We report the metrics computed at the end of every epoch.

\textbf{Pre-train Bert using the C4 dataset:} We train the "bert-base-cased" model of HuggingFace \citep{bertcased} from scratch using the C4 dataset. The model has approximately 110 million trainable parameters. We train the model for 1 million iterations with 10k warm-up steps and a linear decay scheduler. AdamW is trained with a learning rate of $5e-5$ and SGDM is trained with a learning rate of $1e-3$. The weight decay is set to be $0.01$ for both optimizers. Since the training never gets through the whole C4 dataset, we do not reset the value of the tracking quantities. For experiments with constant learning rates, we keep the same configurations but without the scheduler and the warm-up step. We use a batch size of 100000 to compute the large batch loss ($F(x)$) and gradient $\nabla F(x)$. We report the metrics computed every 10k iterations. 

\textbf{Pre-train GPT2 using the Pile dataset:} We train the GPT2 model of HuggingFace \citep{bertcased} from scratch using the Pile dataset. The model has approximately 124 million trainable parameters. We train the model for 1 million iterations with 10k warm-up steps and a linear decay scheduler. Both SGDM and AdamW are trained with a learning rat of $1e-4$. The weight decay is set to be $0.01$ for both optimizers. We do not reset the value of the tracking quantities. For experiments with constant learning rates, we keep the same configurations but without the scheduler and the warm-up step. We use a batch size of 100000 to compute the large batch loss ($F(x)$) and gradient $\nabla F(x)$. We report the metrics computed every 10k iterations. 

\textbf{Testing non-smooth measures:} We train three different tasks with SGDM and AdamW with and without random scaling. We use a variant implementation of SGDM, which updates
\begin{align*}
    \Delta_t = \beta(\Delta_{t-1} - \eta_t g_t), \quad
    x_{t+1} = x_t + s_t\Delta_t.
\end{align*}
$s_t$ is sampled i.i.d. from $\mathrm{Exp}(1)$ with random scaling turned on, and $s_t\equiv 1$ otherwise. 
This is equivalent to SGDM with different effective learning rate and momentum constants, and is shown to have theoretical guarantee \citep{zhang2024random}. We use the standard implementation of AdamW, with the only difference being the inclusion of the additional random scalar.

In the first task, we train the ResNet18 model on the Cifar10 dataset for $200$ epochs with batch size $=128$, with a total of roughly 80k iterations. For SGDM, we use a learning rate $=0.01$ and momentum $\beta=0.99$. For AdamW, we use a learning rate $=3e-4$, weight decay $=0.1$ and default values $b_1=0.9, b_2=0.999$. For both optimizers, we use linear decay scheduler with 5k warmup steps.

In the second task, we train the ``bert-base-cased'' model from scratch on the C4 dataset for 50k iterations with 5k warmup steps and a linear decay scheduler. For SGDM, we use a learning rate $=1e-3$ and momentum $\beta=0.99$. For AdamW, we use a learning rate $=5e-5$, weight decay $=0.01$ and default values $b_1=0.9, b_2=0.999$.

In the third task, we train the GPT2 model from scratch on the Pile dataset for 50k iterations with 5k warmup steps and a linear decay scheduler. For SGDM, weuse a learning rate $=0.01$ and momentum $\beta=0.99$. For AdamW, we use a learning rate $=3e-4$, weight decay $=0.1$ and default values $b_1=0.9, b_2=0.999$. In all tasks, the optimizers with random scaling have the same configuration as its benchmark without random scaling.

\textbf{Runtime:} All experiments are run on 1 NVIDIA
v100 GPUs. Cifar10 experiments take 3 hours, Imagenet experiments take 58 hours, both GPT2 and Bert experiments take about a week to train.

% QZ: I decided to hide the code, since it's no longer anonymous at this point.
% \textbf{Code:} All experiments can be found in the anonymous repository: \url{https://github.com/Neurips24-Submission14212/Submission14212}.

\section{Notations and Definitions}

Below we list all the notations and definitions related to our measurements.

\begin{table}[H]
    \centering
    \begin{tabular}{l l}
        \hline\hline
        Symbol & Description \\
        \hline
        $\instGap_t(\by)$ & Instantaneous convexity gap in iteration $t$ w.r.t. $\by$, defined in \eqref{eq:convexityGap} \\
        $\avgGap_t(\by_{1:t})$ & Unweighted average of $\instGap_i(\by_i)$, defined in \eqref{eq:avg,exp-conv_gap} \\
        $\expGap_t(\by_{1:t})$ & Exponential average of $\instGap_i(\by_i)$, defined in \eqref{eq:avg,exp-conv_gap} \\
        $\convRatio_t$ & Convexity ratio, defined in \eqref{eq:conv_ratio} \\
        $\instSmooth_t(\by)$ & Instantaneous smoothness in iteration $t$ w.r.t. $\by$, defined in \eqref{eq:inst_smooth} \\
        $\expSmooth_t(\by_{1:t})$ & Exponential average of $\instSmooth_i(\by_i)$, defined in \eqref{eq:avg_exp-smooth} \\
        $\maxSmooth_t(\by_{1:t})$ & Maximum over $\instSmooth_i(\by_i)$, defined in \eqref{eq:avg_exp-smooth} \\
        $\gdx_t$ & Update correlation in iteration $t$, defined in \eqref{eq:update_corr} \\
        $\gdelta_t$ & Update correlation with random scaling in iteration $t$, defined in (\ref{eq:update_corr_RS}) \\
        $\dloss_t$ & Instantaneous loss difference in iteration $t$, defined in (\ref{eq:update_corr_RS}) \\
        \hline\hline
    \end{tabular}
    \vspace{1em}
    \caption{Notations of the key identities measured in our experiments.}
    \label{tab:notations}
\end{table}

\section{Extra experiments results}\label{sec:extraExp}
In this section, we report some results that we do not have space to include in the main text.

\subsection{Convexity ratio}
\label{app:convex-ratio}

\begin{figure}[H]
    \centering
    \includegraphics[width=0.5\linewidth]{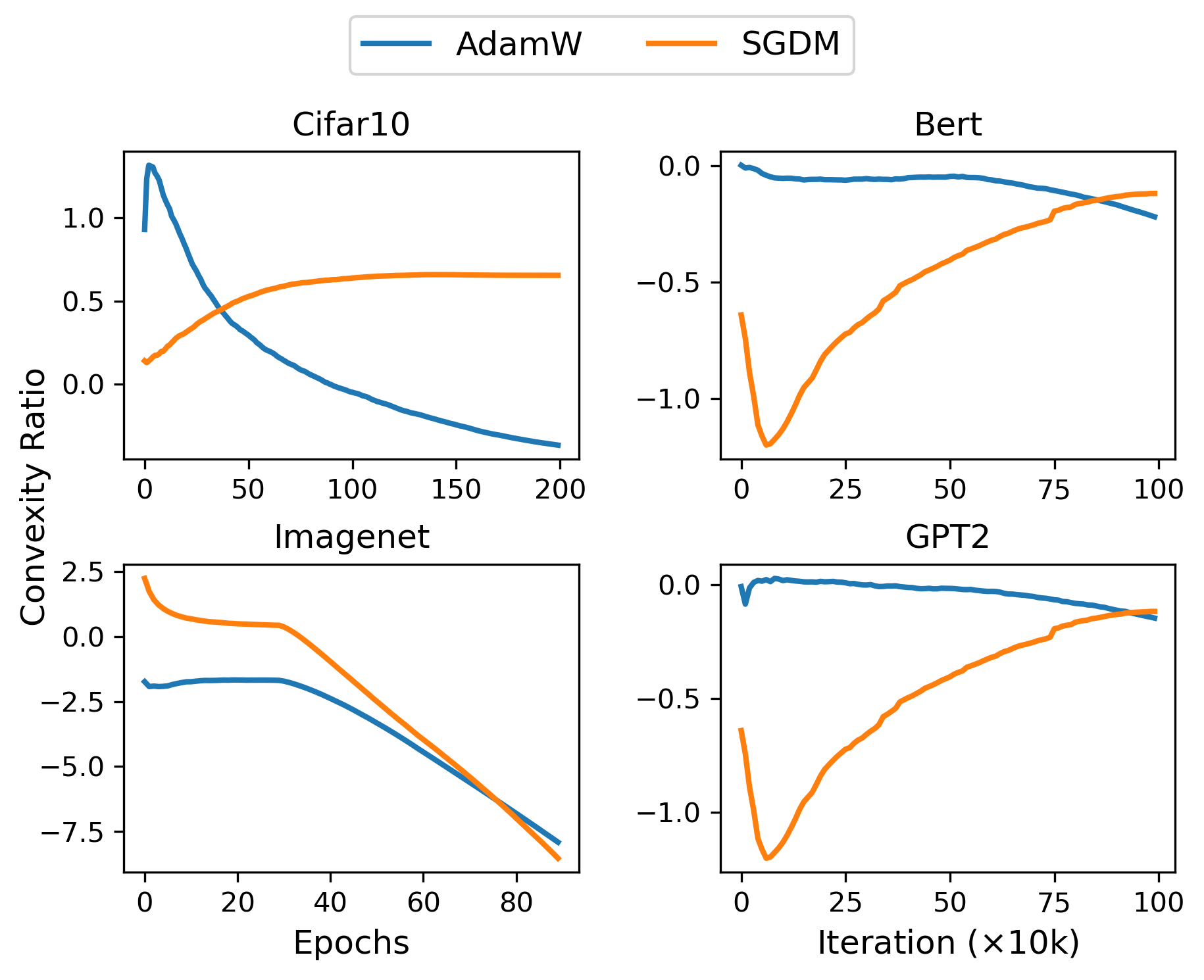}
    \caption{Convexity ratios of deep learning benchmarks where $\bx^\star$ is the final iterate from a \textit{different} training run that has the same configurations but a different random seed. Complementary to Figure \ref{fig:convexityRatio}.}
    \label{fig:convex-ratio-diff}
\end{figure}
In this section, we report the convexity ratio w.r.t the final iterate of a training run with a different seed. As we can see from Fig.\ref{fig:convex-ratio-diff}, the results are almost the opposite of the results in Fig.\ref{app:convex-ratio}. This suggests the existence of multiple ``good'' stationary points. However, the trajectory that the optimizer should take to get to these stationary points are very different and dependent on the initialization. This result also demonstrates the limitations of the classical convexity theory in explaining the success of modern deep learning. Classical convexity analysis suggests that Eq.\ref{eq:conv_ratio} should hold for every $\bx^\star$ but it is not the case in practice.

\subsection{Smoothness with constant learning rates}
\label{app:smoothness}

\begin{figure}[H]
    \centering
    \includegraphics[width=0.5\linewidth]{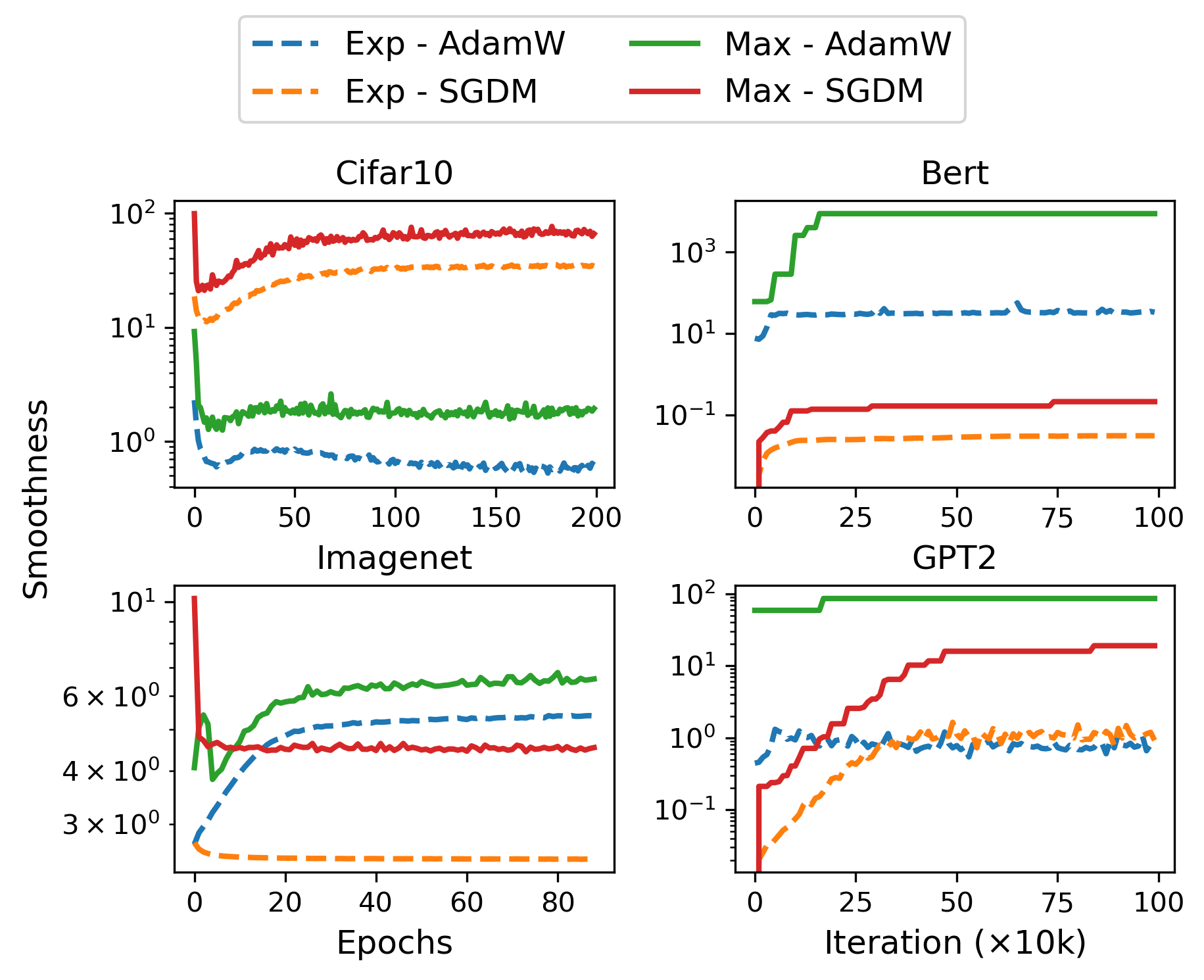}
    \caption{Smoothness measures w.r.t. $\bx_{t-1}$ that uses constant learning rates. Complementary to Figure \ref{fig:smooth}.}
    \label{fig:smoothness-constant-lr}
\end{figure}
To better understand the loss landscapes, we reran all experiments with a constant learning rate (\cref{fig:smoothness-constant-lr}).
With constant learning rates, the loss landscape appeared smoother and more stable. Both the max and exponential average smoothness followed a similar pattern: a rapid drop initially (except for SGDM on ImageNet), followed by a consistent rise until reaching a boundary, then stabilizing. Adam typically achieved smaller (i.e., smoother) measures with a learning rate scheduler, while SGD found smaller measures with a constant rate. We conjecture that this phenomenon suggests that SGD's optimization path is more sensitive to changes in the learning rate, while Adam remains robust across different learning rate settings.

\subsection{The norm of the gradient increases as the training progresses}
\label{subsec:l2}

\begin{figure}[H]
    \centering
    \includegraphics[width=\linewidth]{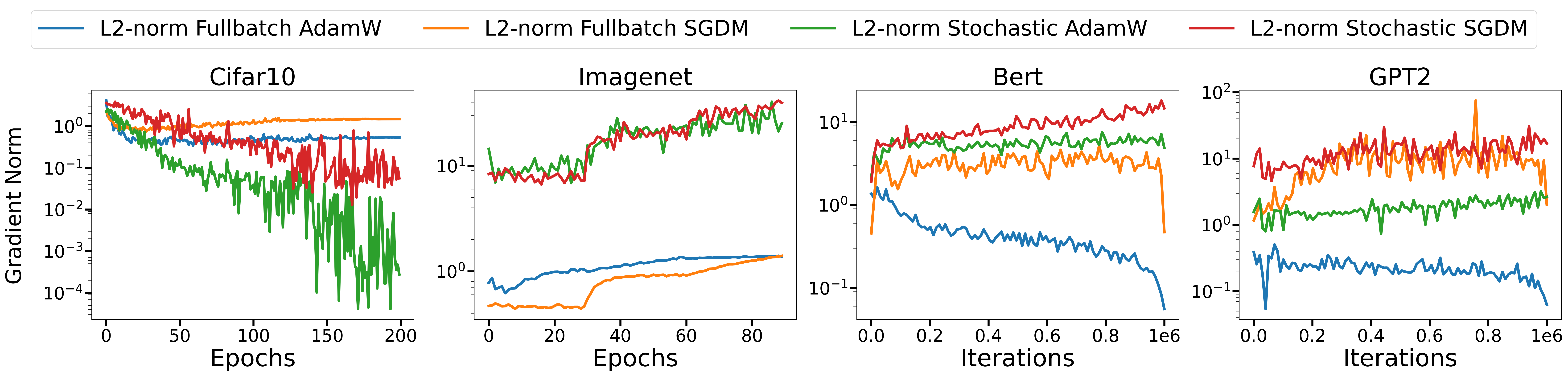}
    \caption{The $L2-$ norm of the gradients as the training progresses.}
    \label{fig:gradnorm}
\end{figure}

When the objective is non-convex, since finding the global minima is NP-hard, previous works focus on finding the $\epsilon-$stationary point \citep{tran2022momentum,fang2018spider,arjevani2020second}, which is defined as a point such that the gradient $\|\nabla F(\cdot)\| \le \epsilon$. The common assumption is that an optimizer performs well if it can find points with a small gradient norm, which is expected to decrease as training progresses. However, as we can see from Fig.\ref{fig:gradnorm}, this is not always the case in practice. In Cifar10 and Bert experiments,the full-batch gradient norms decrease for "good" optimizers (SGDM and AdamW for CIFAR-10, and AdamW for BERT), which supports the theory. Conversely, in the Imagenet and GPT2 experiments, the gradient norms hardly decrease, even though the optimizers are still making consistent progress. In fact, in the Imagenet experiments, the norms actually increase, indicating that we are straying further from the stationary point. This suggests that the use of $\epsilon-$stationary point as the convergence criterion might not be appropriate in practice.

\subsection{Gradient standard deviation increases}

Let us compute the gradient standard deviation as $\sigma \coloneqq \frac{1}{T}\sum_{t=1}^T\|\nabla f(x_t,z_t) - \nabla F(x_t)\|$. Intuitively, the optimizer might make rapid progress if the variance (or standard deviation) is small since it means that our gradient estimate $\nabla f(x_t,z_t)$ is approximating the true gradient well. This is the intuition that leads to the development of a branch of optimization algorithms called variance-reduced algorithms \citep{allen2016variance,cutkosky2019momentum,NIPS2013_ac1dd209}, Thus, we would expect that as the optimizer making progresses, the standard deviation also decreases.

\begin{figure}[H]
    \centering
    \includegraphics[width=\linewidth]{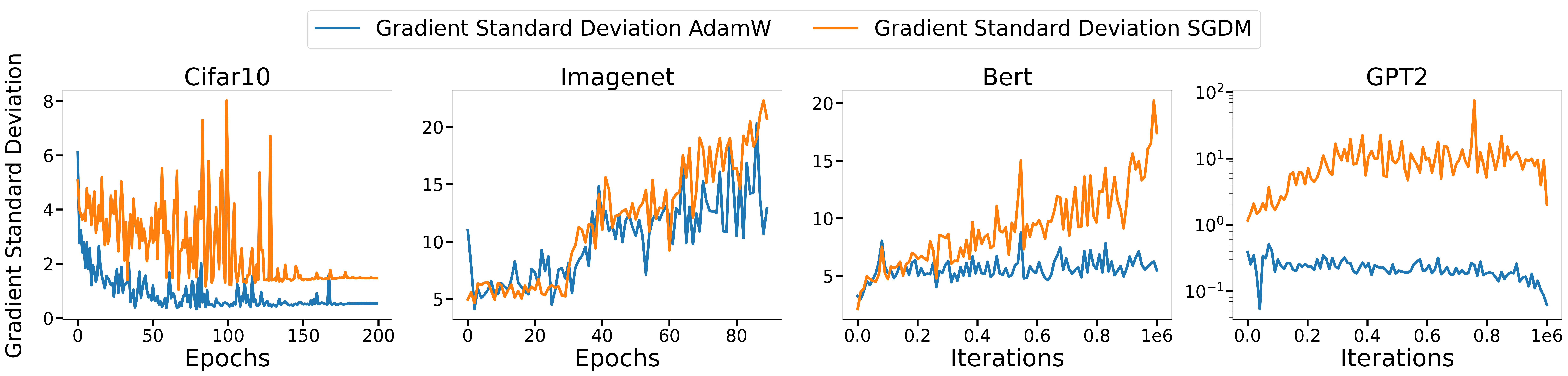}
    \caption{Standard deviation of the gradients}
    \label{fig:std}
\end{figure}

However, similar to the gradient norms, the standard deviation also does not decrease in every experiment. It is hard to conclusively justify why this is the case. One possible explanation for this phenomenon is the existence of multiple minima or low-loss "valley". Thus, even though the optimizer is deviating from the direction to a low-loss "valley" indicating by the true gradient, it is somehow still able to navigate to a different low-loss valley, thus it continues making progress. Further, we note that Adam also consistently returns gradient that is closer to the true gradient. It would be interesting to investigate further to see if this is a property of Adam or of any adaptive method.

\subsection{Parameters norm}

We compute the total parameters norm of the model in each experiment. Adam consistently has larger parameters norm than SGD.

\begin{figure}[H]
    \centering
    \includegraphics[width=\linewidth]{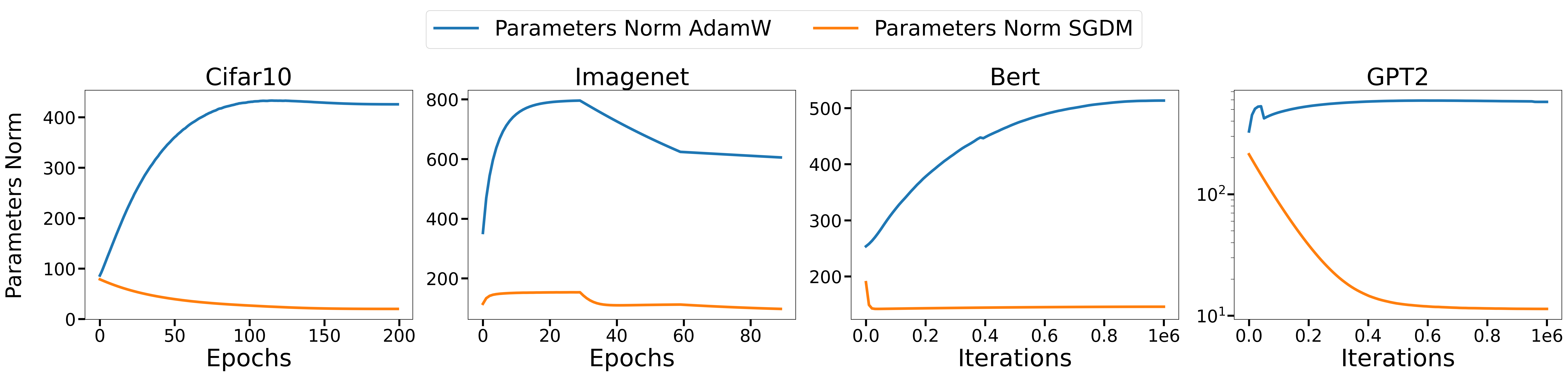}
    \caption{The total $L2-$norm of Model parameters}
    \label{fig:param}
\end{figure}

% Cifar10 and Bert for non-smooth measures
\subsection{$L1-$norm of the stochastic gradients}

\begin{figure}[H]
    \centering
    \includegraphics[width=\linewidth]{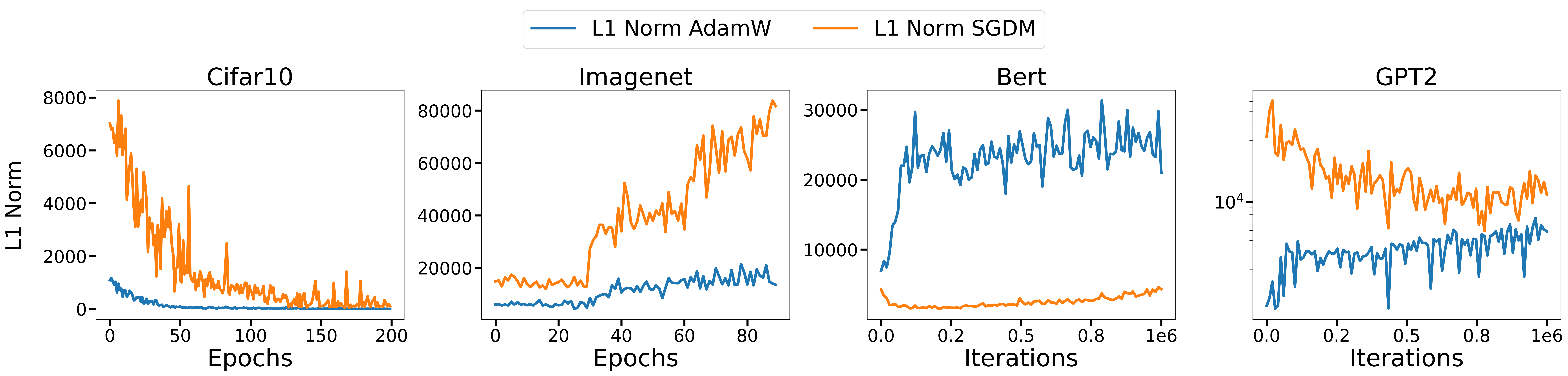}
    \caption{$L1-$norm of the stochastic gradients}
    \label{fig:l1}
\end{figure}

We present additional results in the $L1$-norm of the gradient to complement our $L2$-norm findings discussed in Section \ref{subsec:l2}. An interesting observation is that, although the $L2$-norm of SGD is consistently larger than that of Adam, this is not the case for the $L1$-norm (BERT experiments). This discrepancy suggests that the larger $L2$-norm in SGD may be attributed to outliers in the gradient coordinates, which significantly inflate the final norm. In contrast, Adam, with its adaptive learning rate for each coordinate, effectively minimizes all directions simultaneously, avoiding the issue of gradient outliers.

\subsection{Test accuracy for Image Classification}

\begin{figure}[H]
\centering
 \begin{subfigure}
     \centering
     \includegraphics[width=0.48\textwidth]{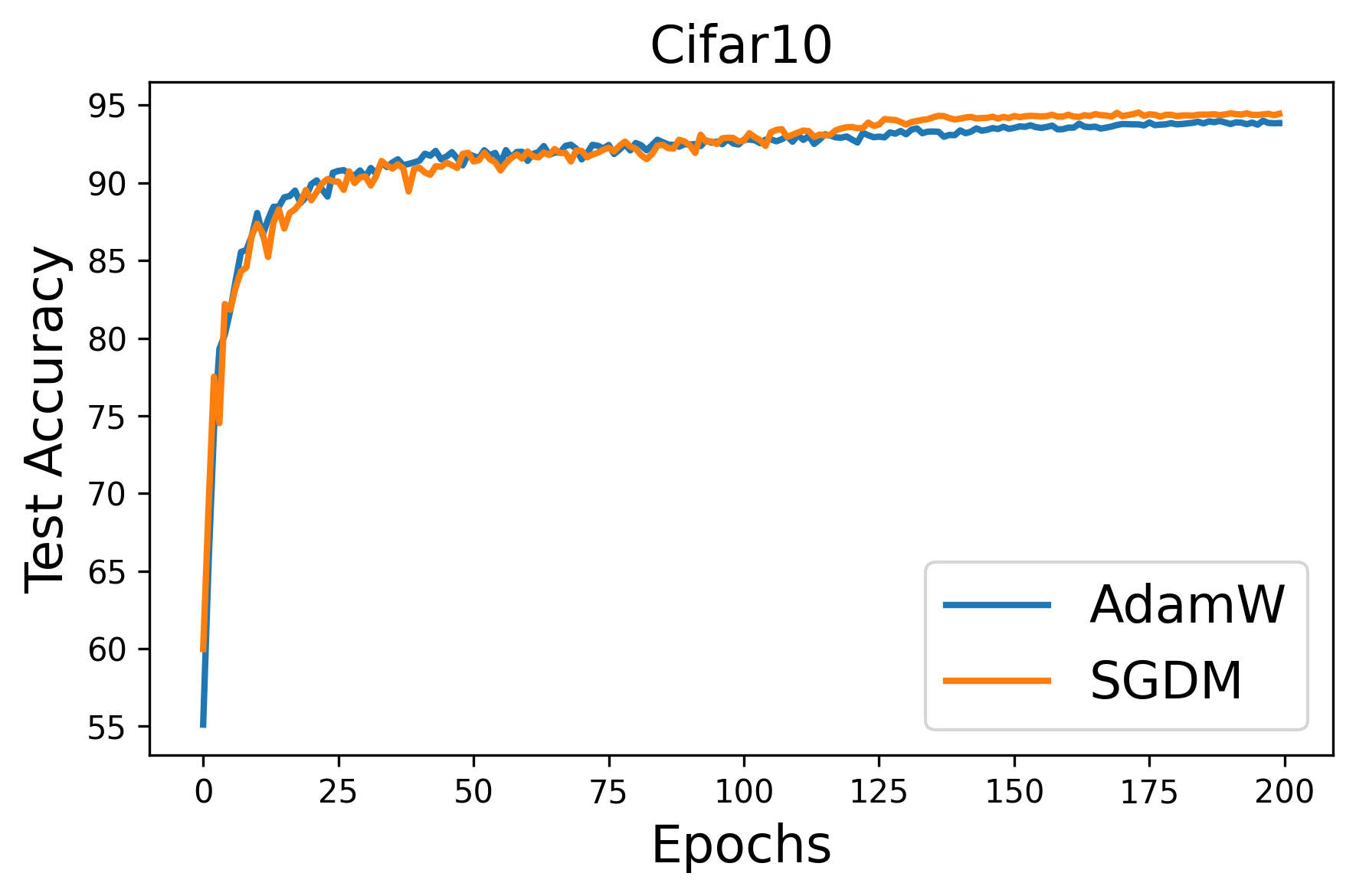}
  \end{subfigure}
  \begin{subfigure}
     \centering
     \includegraphics[width=0.48\textwidth]{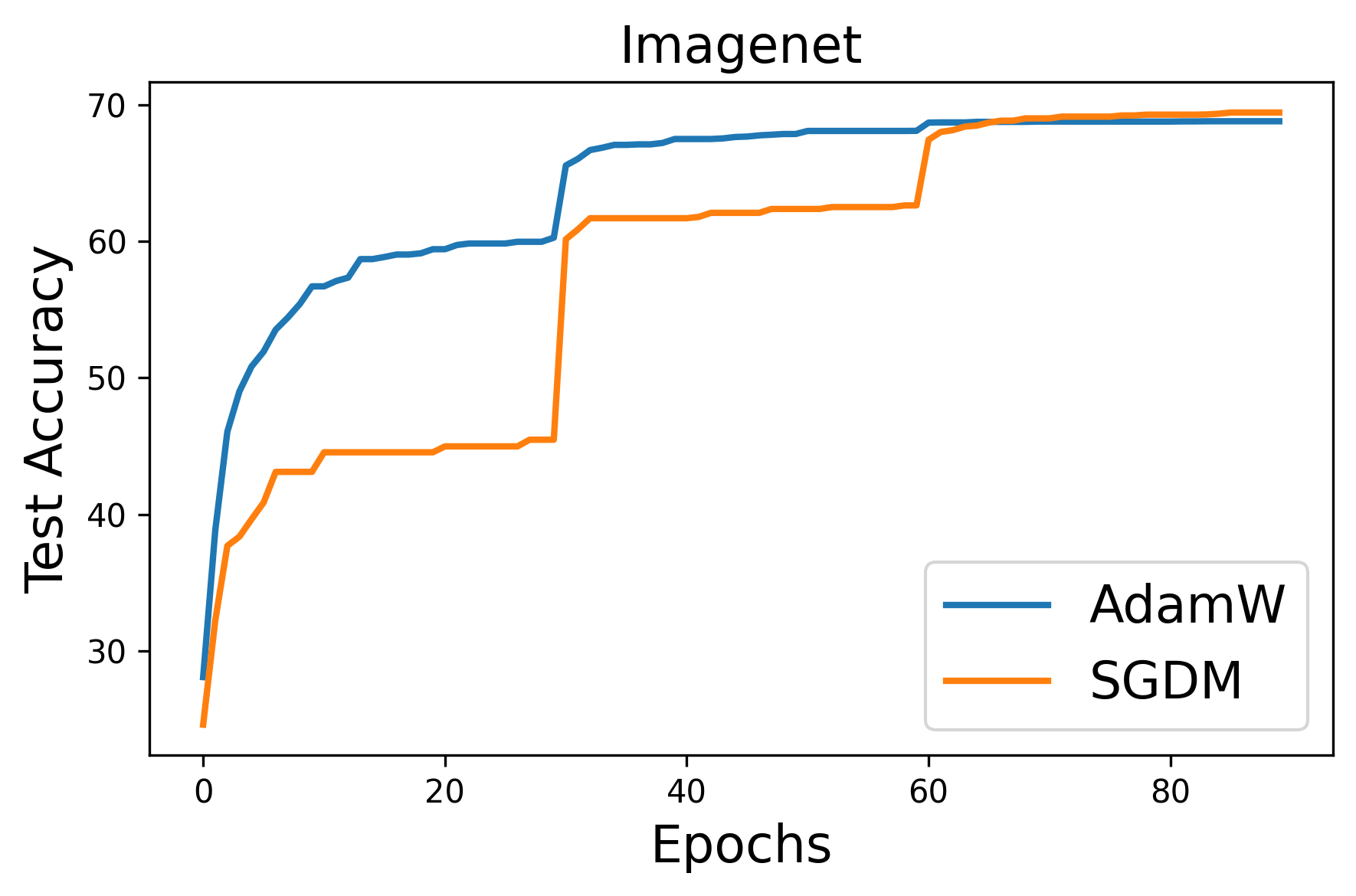}
  \end{subfigure}
    \caption{Test Accuracy of Cifar10 and Imagenet trained on ResNet18}
    \label{fig:test_image_image}
\end{figure}

\subsection{Validation loss of NLP tasks}

\begin{figure}[H]
\centering
 \begin{subfigure}
     \centering
     \includegraphics[width=0.48\textwidth]{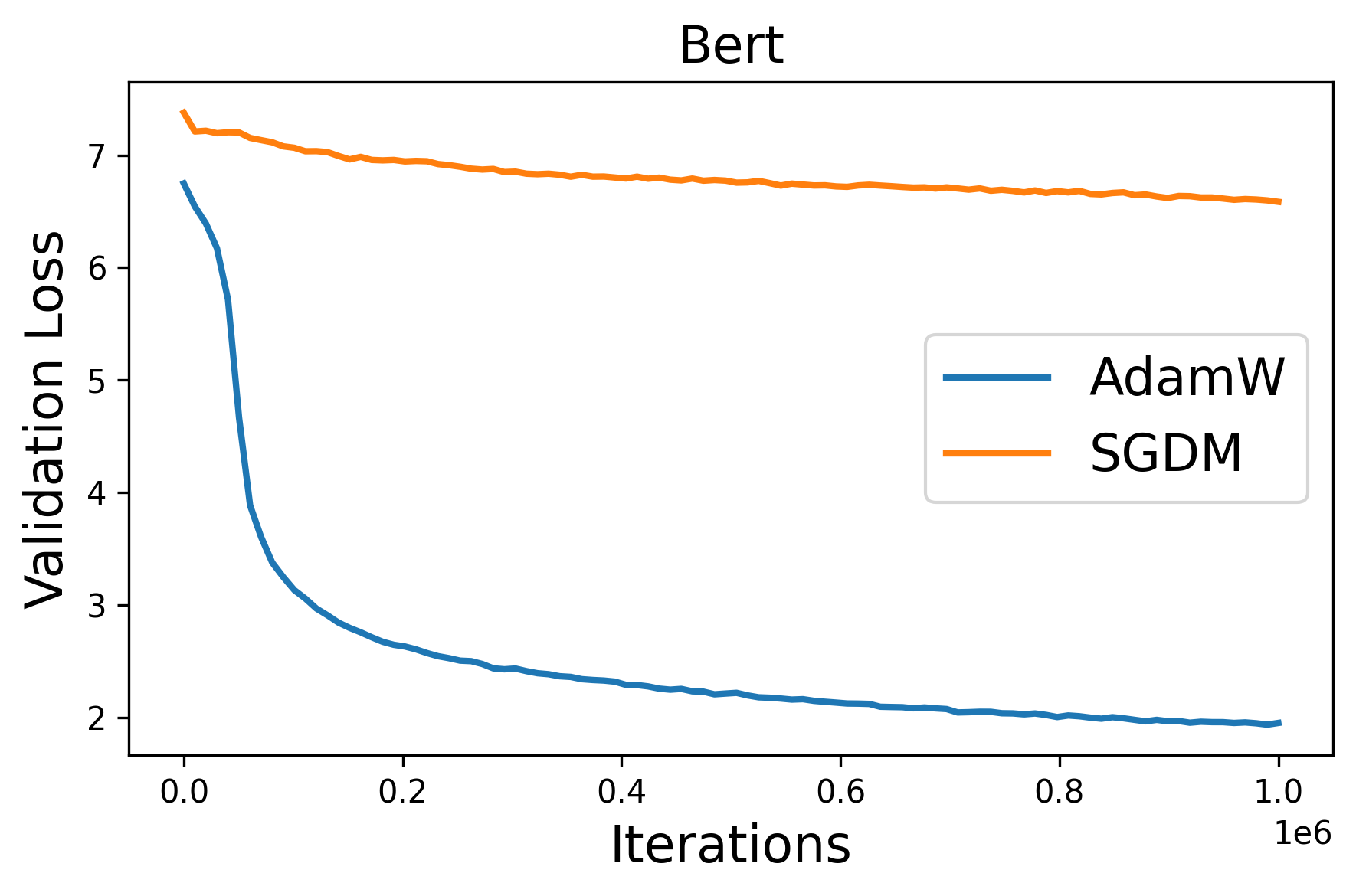}
  \end{subfigure}
  \begin{subfigure}
     \centering
     \includegraphics[width=0.48\textwidth]{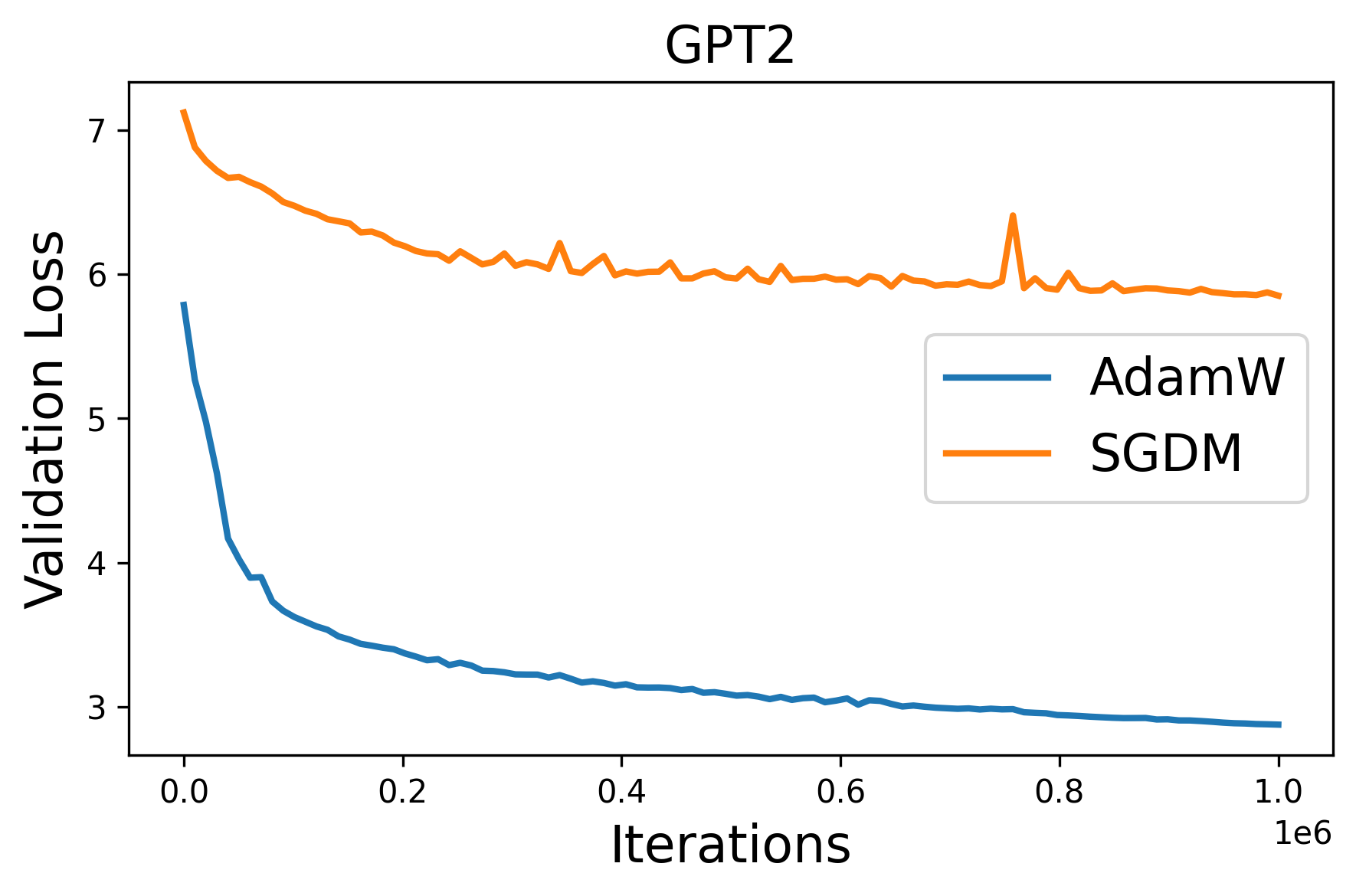}
  \end{subfigure}
    \caption{Validation loss of pre-training Bert on C4 and GPT2 on the Pile}
    \label{fig:test_image_nlp}
\end{figure}

\subsection{Instantaneous convexity gap for deep learning tasks}
\begin{figure}[H]
    \centering
    \includegraphics[width=\linewidth]{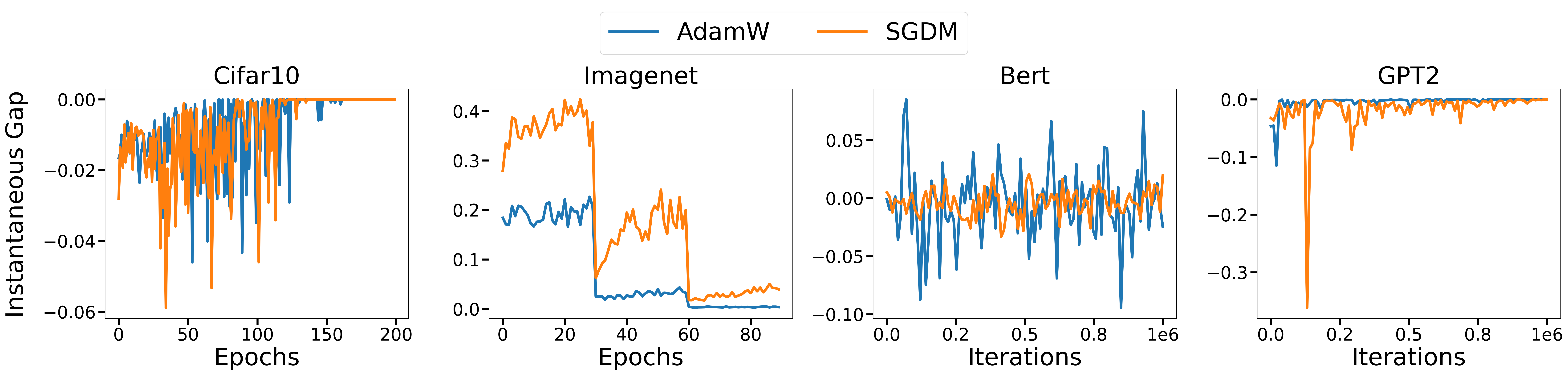}
    \caption{Instantaneous convexity gap w.r.t. $\by_t=\bx_{t-1}$ of deep learning benchmarks. Non-positive gap indicates convexity. See Section \ref{subsec:localconvex} for detailed discussions.}
  \label{fig:instgap}
\end{figure}

\subsection{Update Correlation: Shuffled vs Unshuffled}
\begin{figure}[H]
    \centering
    \includegraphics[width=0.6\linewidth]{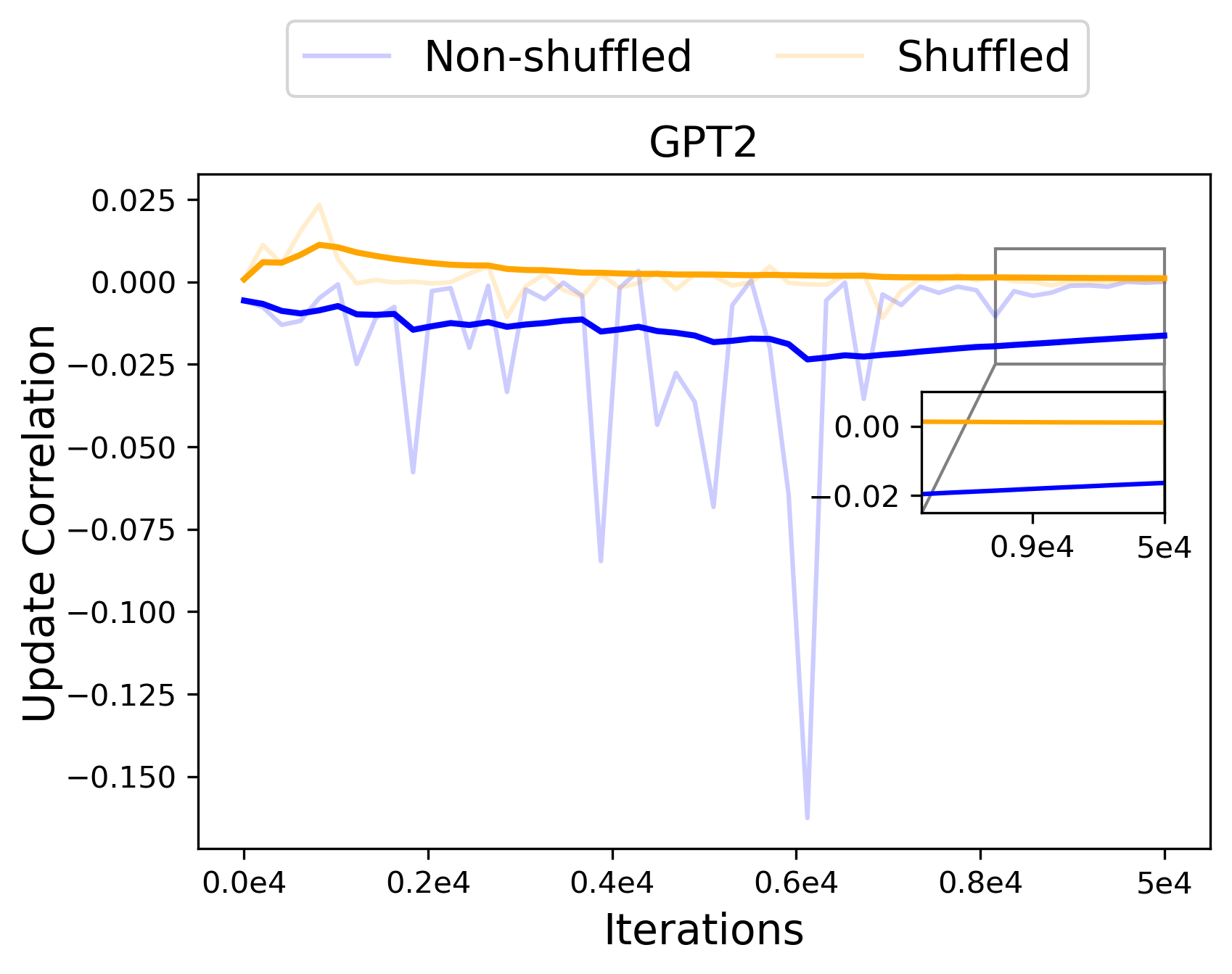}
    \caption{Update correlations of pre-training GPT2 on the Pile dataset - one experiment uses shuffled dataset, the other just iterates through the original dataset. Both are trained for 50k iterations. }
  \label{fig:update_corr_extra}
\end{figure}
\subsection{Non-smooth measures for other deep learning tasks}\label{subsec:nonsmooth}
\begin{figure}[H]
    \centering
    \begin{subfigure}
        \centering
        \includegraphics[width=0.48\textwidth]{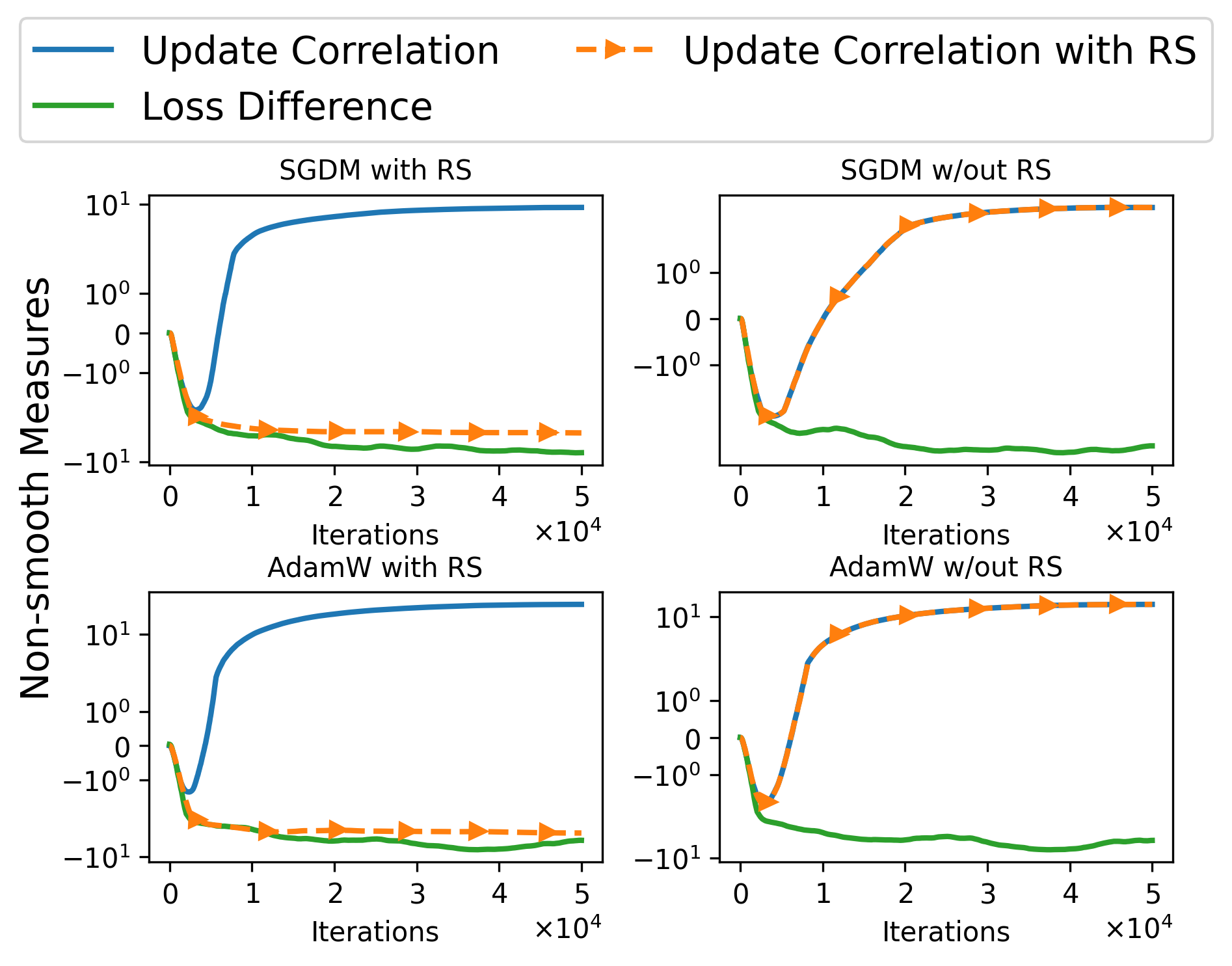}
    \end{subfigure}
    \hfill
    \begin{subfigure}
        \centering
        \includegraphics[width=0.48\textwidth]{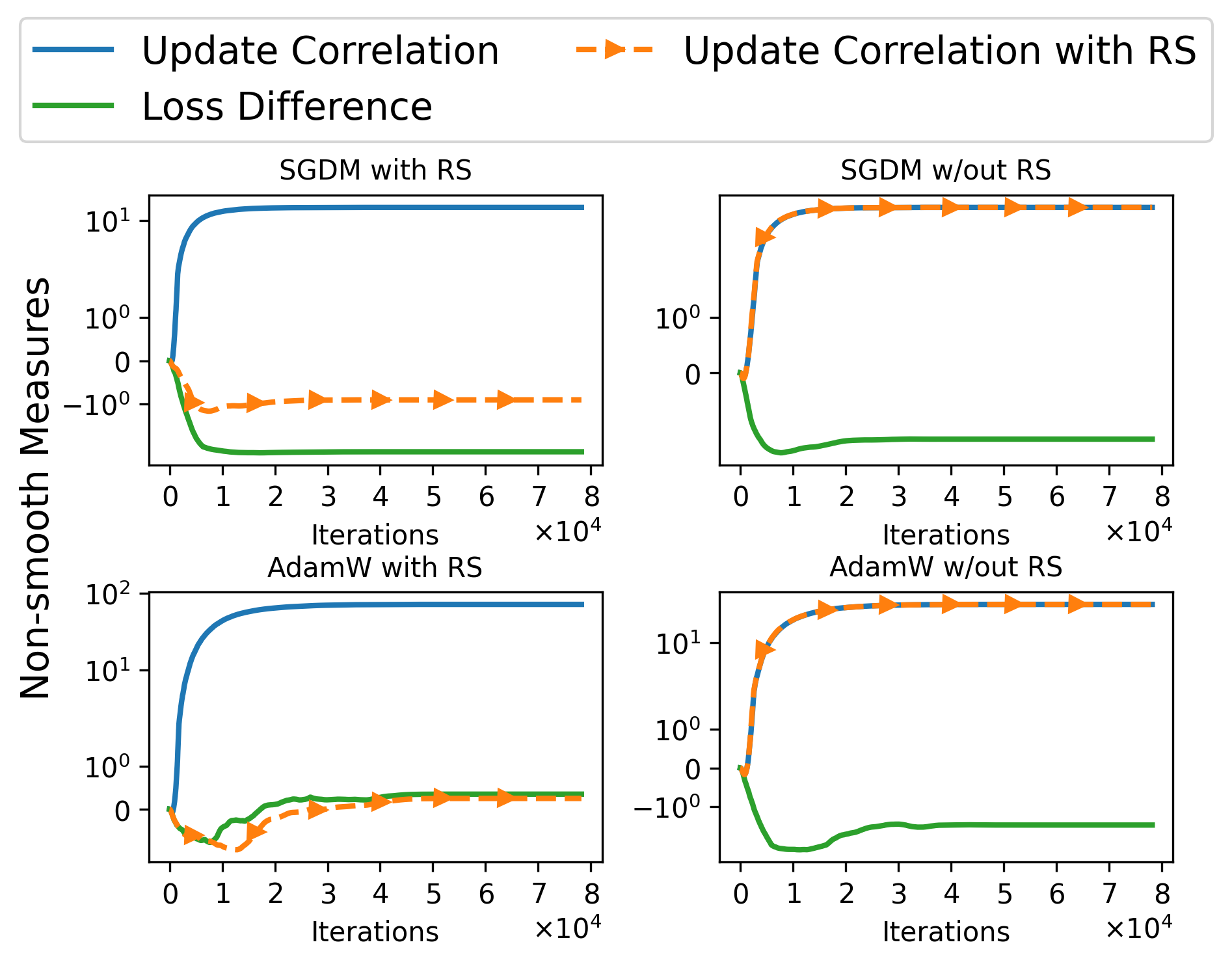}
    \end{subfigure}

    \caption{Cumulative sum (symmetric log scale) of update correlation, update correlation with RS, and loss difference of Bert model trained on C4 dataset (left) and ResNet18 model trained on CIFAR10 dataset (right). Top row is SGDM and bottom row is AdamW; left column is update with RS and right column is the benchmark without RS. See Section \ref{sec:non-smooth} for detailed discussions.}
    \label{fig:non-smooth_bert-renset}
\end{figure}

\end{document}